\documentclass[10pt,twocolumn,letterpaper]{article}

\usepackage{wacv}
\usepackage{times}
\usepackage{epsfig}
\usepackage{graphicx}
\usepackage{amsmath}
\usepackage{amssymb}

\usepackage{algorithmic}
\usepackage{algorithm}
\usepackage{caption}
\usepackage{subcaption}
\usepackage{amsmath}
\usepackage{color, soul}

\usepackage{booktabs}
\usepackage{balance}

\newlength\myindent
\newcommand\bindent{%
	\begingroup
	\setlength{\itemindent}{\myindent}
	\addtolength{\algorithmicindent}{\myindent}
}
\newcommand\eindent{\endgroup}

\wacvfinalcopy 

\def\wacvPaperID{259} 

\ifwacvfinal\fi
\begin{document}

\title{Context-Aware Multipath Networks}

\author{Dumindu Tissera\textsuperscript{*+}, Kumara Kahatapitiya\textsuperscript{*}, Rukshan Wijesinghe\textsuperscript{*+}, Subha Fernando\textsuperscript{+}, Ranga Rodrigo\textsuperscript{*+}\\
\textsuperscript{*}Department of Electronic and Telecommunication Engineering, \textsuperscript{+}CODEGEN QBITS Lab\\
University of Moratuwa\\
\vspace{-0.2in}
{\tt\small \{dumindutissera, kumara0093, rukshandw\}@gmail.com  \{subhaf, ranga\}@uom.lk}}

\maketitle
\ifwacvfinal\thispagestyle{empty}\fi

\begin{abstract}
\vspace{-0.1in}
   Making a single network effectively address diverse contexts---learning the variations within a dataset or multiple datasets---is an intriguing step towards achieving generalized intelligence. Existing approaches of deepening,  widening,  and assembling networks are not cost effective in general. In view of this, networks which can allocate resources according to the context of the input and regulate flow of information across the network are effective. In this paper, we present Context-Aware Multipath Network (CAMNet), a multi-path neural network with data-dependant routing between parallel tensors. We show that our model performs as a generalized model capturing variations in individual datasets and multiple different datasets, both simultaneously and sequentially. CAMNet surpasses the  performance of classification and pixel-labeling tasks in comparison with the  equivalent single-path, multi-path, and deeper single-path networks, considering datasets individually, sequentially, and in combination. The data-dependent routing between tensors in CAMNet enables the model to control the flow of information end-to-end, deciding which resources to be common or domain-specific.
\vspace{-0.1in}
\end{abstract}

\section{Introduction}
\label{se:intro}
\vspace{-0.05in}

\begin{figure}[t]
	\begin{center}
		\begin{subfigure}[]{0.95\linewidth}
			\includegraphics[width=\linewidth]{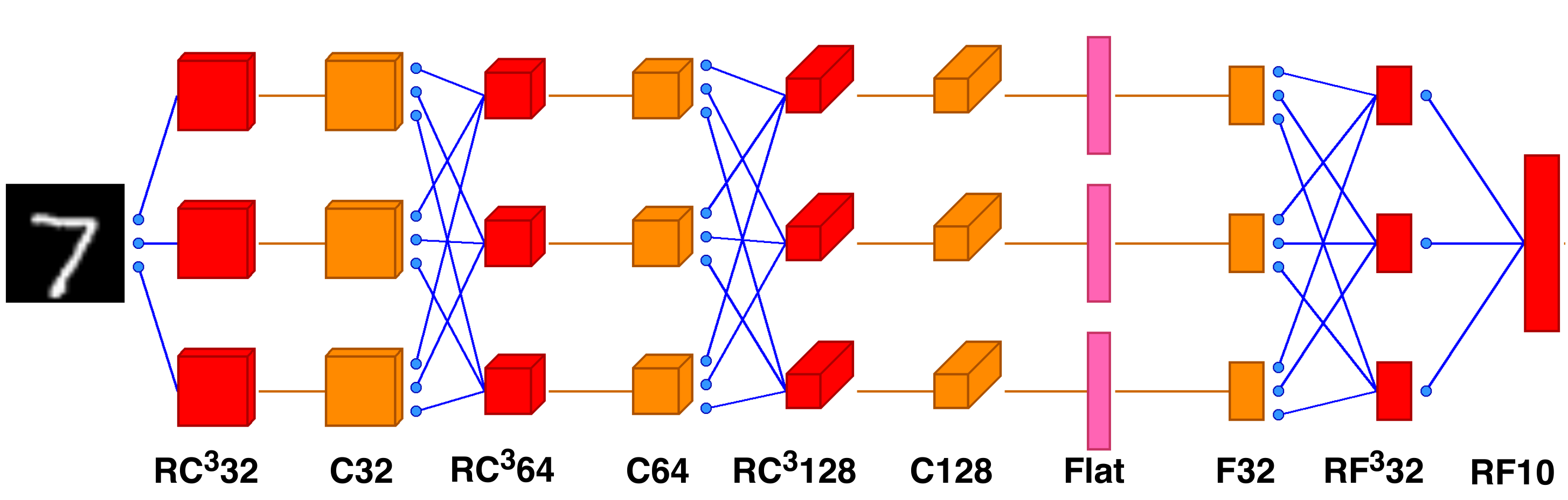} 
			\caption{CAMNet3 Architecture}
			\label{fig:1a}
		\end{subfigure}
		\begin{subfigure}[]{1\linewidth}
			\includegraphics[width=\linewidth]{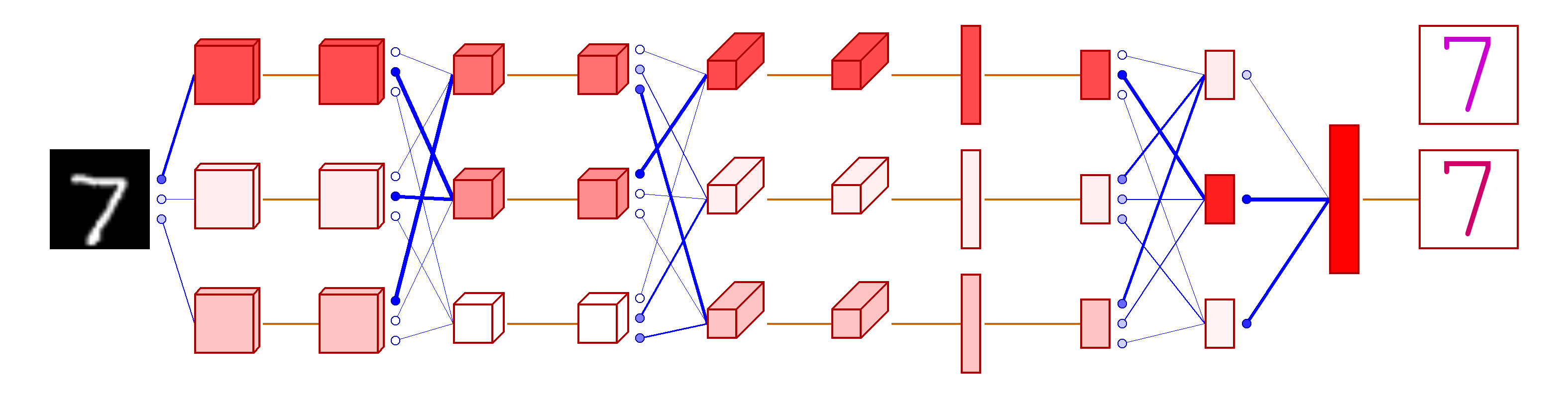} 
			\caption{CAMNet3 on an MNIST digit}
			\label{fig:1b}
		\end{subfigure}
	\end{center}
	\vspace{-0.25in}
	\caption{Illustration of data-dependant routing: Figure \ref{fig:1a} shows CAMNet3 classification architecture: $C$ (convolution) and $F$ (dense) stand for forward layers. $RC$ and $RF$ indicate the presence of routing in these layers. The superscript indicates intended number of tensors in the next layer.  Figure \ref{fig:1b} shows the utilization of this network for a given input in MNIST dataset. Due to learnable gates, CAMNet has data-dependent routing, shown by the color and intensity of routes and tensors (activation paths). }
	\label{fig:camnet_classi}
	\vspace{-0.15in}
\end{figure}

Training a single model to fit data originating from different distributions, or even different datasets altogether, is an interesting problem. This requires the model to adapt based on the context of data. In other words, a model requires to extract different types of features to perform well with different datasets. Thus, it is difficult for a single model, which merely increases the number of  parameters with its depth, to accommodate such variation in feature extraction and perform well in multiple contexts. 

To this end, it is interesting to investigate the idea of a multipath neural network with data dependent soft path assignment,
as such an approach can effectively utilize a large network while sharing resources. Such a single network can learn from varying contexts and even extract richer combination of features to improve performance of a single dataset. 



A model which adapts to variations in the context of the input can provide practical advantages over conventional networks. For instance, (1) the ability to perform well in multiple different datasets of varying contexts concurrently 
and, (2) the ability to adapt to a novel domain while preserving the performance in previously learned domains 
are a few compelling application scenarios to explore.


Approaches towards generalized, intelligent networks that do not merely go deeper receive praise in current literature \cite{ ha2016hypernetworks,  emrouting, sabour2017dynamic, srivastava2015highway}. In addition, learning the mix of shared resources and domain-specific resources is important. 
Adapting to a dataset while preserving already learned information \cite{ewc, li2018learning,piggyback} is also an emerging area of research. Hence, a single network which can generalize to variations in an individual dataset and multiple datasets---in terms of allocating resources in a data-dependant manner---is important.



In this paper, we propose CAMNet, a multi-path convolutional neural network, which employs a data-dependent routing mechanism to route among parallel tensors. Such a learnable routing mechanism helps regulate the flow of information resulting in rich feature extraction and better utilization of resources at runtime (Fig. \ref{fig:camnet_classi}). Thus, the novelty of CAMNet is twofold:
\begin{itemize}
	\vspace{-0.08in}
	\item Multiple parallel tensors in each layer which extract different sets of features, 
	accommodating better feature utilization for different domains.
	\vspace{-0.1in}	
	\item A novel learnable, data-dependant routing algorithm which controls the data flow among parallel tensors in subsequent layers. This allows the network to select the data path based on the context of the input image, resulting in a single model which performs well in different contexts.
\end{itemize}

In the proposed routing algorithm of CAMNet, each tensor among the parallel set of tensors in each layer predicts each tensor in the succeeding layer, which is made data-dependent with the introduction of learnable parameters in this prediction. 
Our intuition is that extracted features should be dependent upon the context of the input, in the case of learning from diverse data. 

To evaluate the effectiveness of CAMNet, we first show the performance gain in individual classification datasets by CAMNet, followed by jointly optimizing CAMNet on a combined dataset of multiple domains, which shows variation in activation paths and better overall performance. Then, we evaluate how well CAMNet adapts to new domains, preserving the performance in previously-learned contexts, considering the Learning without Forgetting (LwF) technique \cite{li2018learning}. We show the variation of path combinations taken by CAMNet for different domains. Finally, we empirically show that different parallel paths have learned different weight distributions.

The rest of the paper is organized as follows. We present related literature in Section \ref{se:relatedwork} and formulate the network architecture along with the routing algorithm in Section \ref{se:systemarchitecture}. In Section \ref{se:experiments}, we show the performance of CAMNet on individual and joint datasets as well as lifelong learning. 
Finally, we discuss the outcomes and potential further extensions of CAMNet in Section \ref{se:discussion} and conclude in Section \ref{se:conclusion}.

\section{Related Work}
\label{se:relatedwork}
\vspace{-0.05in}
Having deep networks for generalizing to wide-range datasets is a common practice
. Although the depth influences the performance of a neural network, training such networks is difficult and the general intelligence of this kind of a network is questionable \cite{srivastava2015highway}. Hence, more attention is paid towards generally intelligent neural networks. 

One such approach is to harvest more information within a layer in a neural network. Capsule networks \cite{emrouting,sabour2017dynamic} involves extracting information about pose and orientation where, instead of convolutional scalars, there are vectors and matrices as layer outputs. 

An alternate approach towards general intelligence is to build more flexible networks where the network is supposed to learn more than the weights. Hypernetworks \cite{ha2016hypernetworks} includes a smaller network embedded inside the large network where the small network is predicting the weights of the large network. Squeeze-and-excitation networks \cite{hu2017squeeze} introduced learnable weighting of each convolutional channel in order to build a generalized model across different datasets.  Highway networks \cite{srivastava2015highway} proposed the use of gates which learn to regulate the flow of information across the network which enabled deep models to be trained effectively. However, these approaches use a single-path network. Although the weights might vary, each training or testing input is fed through a single path. In contrast, our model is able to benefit from parallel paths with different weights on each path.


The term multitask learning is defined in several ways depending on the nature of the application. Sharing information between similar tasks \cite{donahue2014decaf, pentina2015curriculum}, performing multiple tasks on a single input \cite{nddr-cnn,fully-adaptive-mtl, mtl-soft-layer-ordering, cross_stich, routing-networks, sluice,  wang2015designing,pad-net,   zhang2014facial}, and learning from multiple domains \cite{kang2011learning, li2018learning, packnet, residual-adapters} can be considered as few such applications where the multitasking is defined in different forms. However, our work highlights the capability of CAMNet to perform context awareness in individual datasets and to learn from multiple domains either sequentially or simultaneously. 

The common multitasking (MTL) in computer vision refers to processing multiple different tasks in a single input (e.g., semantic segmentation and surface-normal prediction). Conventional approaches to MTL includes using  shared layers to some extent along with  task-specific layers. Choosing the number of task-specific layers and shared layers is task dependant. However, recent approaches solve the problem of choosing from possible combinations in this context by letting the model to learn the use of shared and task-specific layers according to the task. Cross-stitch networks \cite{cross_stich} and sluice networks \cite{sluice} introduce sharing resources between parallel networks where communication between parallel layers is done through learning a linear combination of parallel tensors. NDDR-CNN \cite{nddr-cnn} use discriminative dimensionality reduction to fuse features from parallel tensors. 

However, our intention in this work is to build an intelligent model which can handle a diversity in a dataset, especially in input domain, to an extent where the model can even handle multiple datasets at a time. The main difference between CAMNet and these approaches is that our model learns the routing between parallel tensors in a data-dependant manner, starting from the input. Also, all the parallel paths share the same single input and output making CAMNet a single-input-single-output model. A routing layer learns from data while forwarding the incoming flow to particular set of tensors in the next layer.

Lifelong learning involves learning from multiple datasets one after the other. Conventional approaches include fine tuning \cite{fine_tuning} and feature extraction \cite{donahue2014decaf}
which suffer from catastrophic forgetting. Rebuffi \etal \cite{rebuffi2017icarl} introduced incremental classification as opposed to batch training in order to overcome catastrophic forgetting. Learning without Forgetting (LwF) \cite{li2018learning} and Elastic Weight Consolidation (EWC) \cite{ewc} are also two approaches introduced to overcome this issue in terms of modifying the objective function. In contrast, PackNet \cite{packnet} and Piggyback \cite{piggyback} methods use binary masking on dense weight filters once trained in a dataset in order to free up the least used weights to learn from the next dataset. However, these approaches need larger filters depending on the number of datasets to be trained sequentially. 



Approaches that can gradually build customized networks according to the input are also inspiring for our research. ConvNet-AIG \cite{convnet-aig} and BlockDrop \cite{blockdrop} are two approaches introduced for data dependant choosing of residual blocks in a deep network as alternatives to conventional Residual Networks \cite{resnet}. These approaches learn which residual block to keep according to the nature of the input.

The idea of ensembles is quite popular where multiple neural networks are stacked together to enhance the cumulative performance on a single dataset. However, deploying a complex ensemble model for learning from diverse data would be challenging and computationally expensive in the presence of complex deeper networks as individual models \cite{bucilua2006model}. There is also the computational redundancy where multiple networks learning similar patterns and weights. Also, it has been shown that the knowledge gained by an ensemble model can be compressed in to a single model which is much simpler \cite{hinton2015distilling}. Our approach differs from ensembles in the network architecture itself where our network is not a stack of different paths. Our network is able to learn from different datasets simultaneously while sharing intermediate tensors.




\section{Architecture and Method}
\label{se:systemarchitecture}
\vspace{-0.05in}

As the intuition behind CAMNet is to design a context-aware network which can differentiate subtle variations within a dataset, or even between multiple different datasets and learn sequentially with minimal changes to previously learned features, we consider a model with parallel paths. In contrast to conventional networks with a single path of data flow (series of tensors) subjected to convolution and fully connected operations, CAMNet consists of multiple parallel data paths as such. The proposed data-dependent and learnable routing algorithm delegates the data flow from the input among these parallel paths and recombine to produce the single output of the network. It routes between subsequent layers based on two main operations: the prediction and the construction. The prediction operation involves each tensor in a certain layer predicting each tensor in the subsequent layer and the construction operation computes each tensor in the subsequent layer utilizing the predictions made by all the tensors from the previous layer for that particular tensor.

\subsection{Routing between Tensors}
\label{ss:_sa_routing}
\vspace{-0.05in}

\begin{figure}[t]
	\begin{center}
		\includegraphics[width=0.7\linewidth]{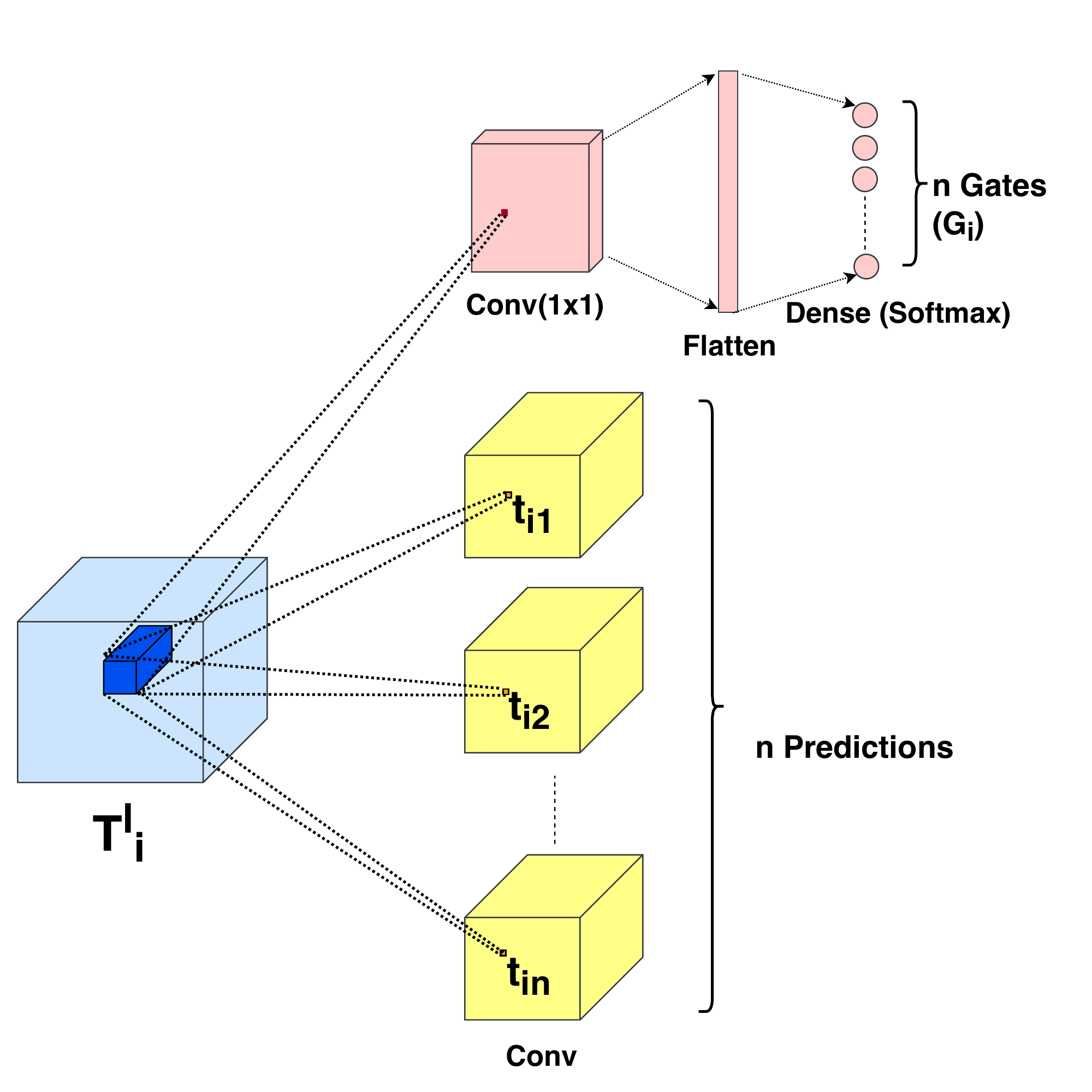}
	\end{center}
	\vspace{-0.2in}
	\caption{Operations carried out by a 3-dimensional tensor $i$ in layer $l$ ($T^{l}_{i}$). $1\times1$ convolution limits the amount of parameters fed into the dense gate operation.}
	\label{fig:block}
	\vspace{-0.1in}
\end{figure}

Suppose we have $m$ tensors in a particular layer ($l$), and $n$ tensors in the subsequent layer ($l+1$). We denote tensor $k$ of layer $l$ as $\boldsymbol{T}^{l}_{k}$. 
Each tensor in layer $l$ performs a prediction for each tensor in the layer $l+1$. The prediction $\boldsymbol{t}_{ij}$ made by tensor $i$ in layer $l$ to tensor $j$ in layer $l+1$ is given by, 
\vspace{-0.1in}
\begin{equation}
\boldsymbol{t}_{ij} = W_{ij}\boldsymbol{T}^l_{i} + b_{ij},
\end{equation} 

where $ W_{ij}$ and $b_{ij}$ correspond to weight and bias terms, respectively. In implementation if $\boldsymbol{T}^l_i$ is a 3-dimensional tensor, this operation is carried out via a standard convolution (Fig \ref{fig:block}). Moreover, each tensor in layer $l$  predicts $n$ dimensional vector of gate values: $\boldsymbol{G}_i$, which represent the probabilities of the corresponding tensor $i$ being routed to each tensor in layer $l+1$. \vspace{-0.1in}
\[\boldsymbol{G}_i = [g_{i1}, g_{i2}, \dots, g_{in}]  \] 
Here, $g_{ij}$ corresponds to the scalar gate value connecting $i^{th}$ tensor in $l^{th}$ layer ($T^l_i$) to  $j^{th}$ tensor in layer $l+1$ ($T^{l+1}_j$). $\boldsymbol{G}_i$ is given by,
\vspace{-0.1in}
\begin{equation}
\boldsymbol{G}_i = \mathrm{softmax}(W_{G_{i}}\boldsymbol{T}^{l}_{i} + b_{G_{i}}),
\label{eq:gate_prob_computation}
\end{equation}

where $ W_{G_{i}}$ and $b_{G_{i}}$ correspond to weight and bias terms.
In practice, if $\boldsymbol{T}^l_i$ is a 3-dimensional tensor, we first run a $1\times1$ convolution before applying the dense operation in Eq. \ref{eq:gate_prob_computation} to reduce the number of parameters. The activation function $\mathrm{softmax}(.)$ limits the utilization of multiple parallel tensors for a certain context so that each tensor is more likely to be allocated to a single tensor in the subsequent layer. Figure \ref{fig:block} shows the operations carried out by a 3-dimensional tensor in layer $l$ in the prediction phase.


The tensors in layer $l+1$ are calculated based on the predictions and gate values provided by layer $l$. The predictions are weighted by the corresponding gate and a non-linear activation is applied to produce each tensor in layer $l+1$.
Eq. \ref{eq:tensor_computation} shows how the of tensor $j$ in layer $l+1$ ($\boldsymbol{T}^{l+1}_{j}$) is calculated using the predictions made by tensors in layer $l$ ($\boldsymbol{t}_{ij}$, $i=1, 2, \dots,  m$) and the gates connecting corresponding tensors in layer $l$ to the particular tensor in layer $l+1$ ($g_{ij}$, $i=1, 2 \dots, m$).
\vspace{-0.1in}
\begin{equation}
\boldsymbol{T}^{l+1}_{j} = f\left(\sum_{i=1}^{n} (g_{ij}\times \boldsymbol{t}_{ij})\right).
\label{eq:tensor_computation}
\end{equation}

\begin{figure}[t]
	\begin{center}
		\includegraphics[width=0.8\linewidth]{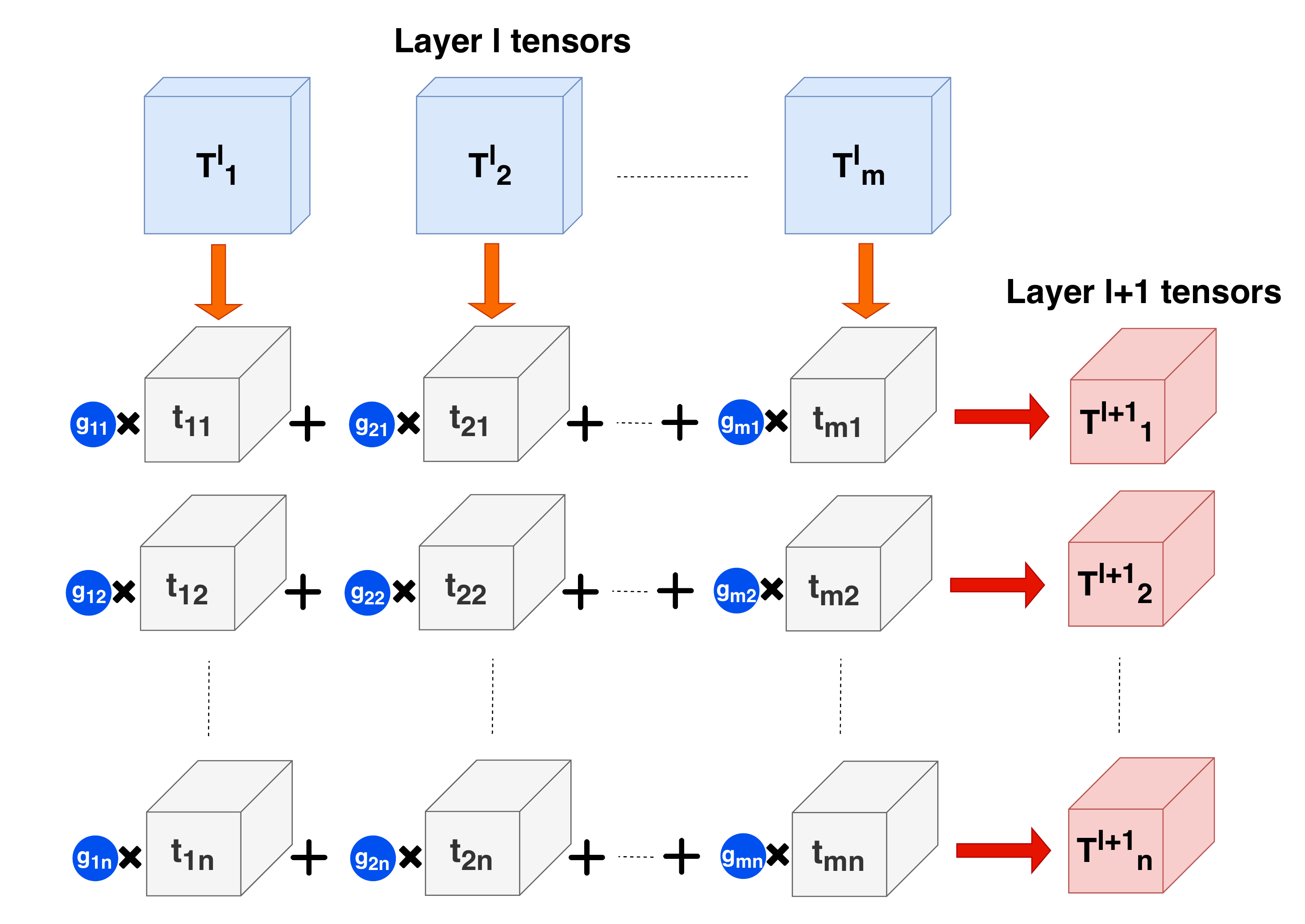}
	\end{center}
	\vspace{-0.2in}
	\caption{Constructing layer $l+1$ based on the predictions and gates computed by layer $l$ See Eq. \ref{eq:tensor_computation}}
	\label{fig:routing}
	\vspace{-0.1in}
\end{figure}

Here, $f$ stands for the non-linear activation function applied to the weighted prediction. We use $\mathrm{ReLU}(.)$ in the intermediate layers and either $\mathrm{softmax}(.)$ for classification or $\mathrm{tanh}(.)$ for image-to-image translation in the final layer. Figure \ref{fig:routing} shows the construction of layer $l+1$ from layer $l$ predictions and gates.
Algorithm \ref{alg:routing} demonstrates the routing algorithm between subsequent layers.

\begin{algorithm}[ht]
	\caption{Prediction and routing algorithm between layer $l$ and $l+1$}
	\label{alg:routing}
	\begin{algorithmic}
		\STATE {\bfseries Input:} 
		\bindent
		\STATE $\boldsymbol{T}^{l}$: layer $l$, \COMMENT{$\boldsymbol{T}^l$ = [$\boldsymbol{T}^l_i$ for $i=1,2,\dots,m$]}
		\STATE $m$: number of tensors in layer $l$ 
		\STATE $n$: number of tensors in layer $l+1$  
		\eindent
		
		\STATE {\bfseries Predictions from current layer:}
		\bindent
		\FOR{$i=1$ {\bfseries to} $m$}
		\FOR{$j=1$ {\bfseries to} $n$}
		\STATE $\boldsymbol{t}_{ij} \leftarrow W_{ij}\boldsymbol{T}^{l}_{i} + b_{ij}$ 
		\ENDFOR
		\STATE $\boldsymbol{G}_i \leftarrow \mathrm{softmax}(W_{G_{i}}\boldsymbol{T}^{l}_{i} + b_{G_{i}})$ 
		\ENDFOR	
		\eindent	
		
		\STATE {\bfseries Construction of next layer:}
		\bindent
		\FOR{$j=1$ {\bfseries to} $n$}
		\STATE $\boldsymbol{T}^{l+1}_{j} \leftarrow  f(\sum_{i=1}^{n} (g_{ij}\times \boldsymbol{\mathit{t}}_{ij}))$
		\ENDFOR
		\eindent	
		
		\STATE {\bfseries Return:} 
		\bindent
		\STATE $\boldsymbol{T}^{l+1}$:layer$l+1$,
		\COMMENT{$\boldsymbol{T}^{l+1}$=[$\boldsymbol{T}^{l+1}_j$ for $j=1,2,\dots,n$]}
		\eindent
	\end{algorithmic}	
\end{algorithm}

\subsection{Forward Layers}
\label{ss:sa_forw_layers}
\vspace{-0.05in}
We add layers which do not perform routing, to deepen the network without increasing the computational overhead due to additional routing layers. We refer to such layers as \textit{forward layers}, which can be either convolutional or fully-connected layers (one or several following each tensor) as depicted in Fig. \ref{fig:1a} (e.g., $C32$ in orange). Forward layers help make CAMNet deeper and isolate the learning process of individual paths for a single step.

\subsection{Training}
\label{ss:sa_training}
\vspace{-0.05in}
Since the routing layers incur convolutional or dense predictions, initializing CAMNet with pretrained single path neural networks is impractical. Moreover, stitching pretrained models trained on different tasks does not align with the intuition of this work. Thus, all the training is done from scratch. 

We perform lifelong learning on different datasets with the use of Learning without Forgetting (LwF) algorithm \cite{li2018learning}. LwF is based on the principle of assigning new task-specific parameters ($\theta_n$) before training on a subsequent dataset, keeping the previous task-specific layer intact ($\theta_o$). The loss function is a weighted loss of the cross-entropy for the new task and the knowledge distillation loss \cite{hinton2015distilling} for the old task.







\section{Experiments}
\label{se:experiments}
\vspace{-0.05in}
We conduct various experiments to show the ability of our model to generalize in multiple datasets separately, accommodating the variation within a dataset and in multiple different datasets combined. Moreover, we illustrate the effectiveness of the data-dependant routing in contrast to the equivalent multi-path network without connections between parallel tensors. We vary the number of parallel tensors to investigate the effect of the width of CAMNet on the performance. For these evaluations, we consider classification and pixel-labeling as test cases.

\begin{table}[ht]
	\caption{Compared Networks: $C$ denotes a convolutional layer and $F$ stands for a fully connected layer. Suffix $r$ stands for a routing layer. CAMNetX stands for the CAMNet version carrying X number of tensors in each layer.}
	\vspace{-0.2in}
	\label{tab:compared_networks}
	\begin{center}
		\renewcommand{\tabcolsep}{0.5mm}
		\begin{tabular}{@{}lr@{}}
			\toprule
			Network & Structure \\
			\midrule
			\small BaseCNN & \footnotesize $C32$ $C32$ $C64$ $C64$ $C128$ $C128$ $F32$ $F32$ $F10$ \\
			\small BaseCNN2 & \footnotesize $C32$ $C32$ $C64$ $C64$ $C128$ $C128$ $C128$  \\& \footnotesize  $C256$  $C256$ $C256$ $F32$ $F32$ $F10$ \\
			\small MultiCNNX & \small X number of parallel BaseCNNs  \\
			\midrule
			\small CAMNetX & \footnotesize $^rC32$ $C32$ $^rC64$ $C64$ $^rC128$ $C128$ $F32$ $^rF32$ $^rF10$ \\& \small (Equivalent single path BaseCNN) \\
			\small tinyCAMNetX & \footnotesize $^rC16$ $C16$ $^rC32$ $C32$ $^rC64$ $C64$ $^rF10$ \\
			\midrule
			\small SENet & \footnotesize Equi. to BaseCNN, All convolutions with SE operation \\
			\small Cr-Stitch2 & \footnotesize Equi. to CAMNet2 with stitching replaced routing  \\
			\bottomrule
		\end{tabular}
	\end{center}
	\vspace{-0.2in}
\end{table}

\subsection{Datasets}
\label{ss:ex_datasets}
\vspace{-0.05in}
For classification, we use MNIST~\cite{mnist}, Fashion MNIST~\cite{fashion}, KMNIST~\cite{kmnist}, NotMNIST~\cite{notmnist}, CIFAR10 \cite{cifar10} and SVHN \cite{svhn} datasets.  For pixel-labeling, we use a few existing datasets on image-to-image translation and semantic segmentation. Furthermore, we consider the CMP facades dataset \cite{Tylecek13}, satellite to aerial map dataset extracted from Google maps and used in pix2pix \cite{isola2017image}, Cityscapes dataset \cite{Cordts2016Cityscapes} and KITTI road segmentation dataset \cite{Fritsch2013ITSC}.  

In addition to these datasets, we present our own dataset of drone-image to road-topology translation, which contains around 300 annotated image pairs. These images are sampled from several videos of drones following roads, hence, contain a considerable variety in the context of the input. 
Here, we only annotate the road and the background, disregarding other objects, in contrast to road-segmentation datasets.

\subsection{Compared Architectures}
\label{ss:compared_archi}
\vspace{-0.05in}
Table \ref{tab:compared_networks} illustrates the baseline networks and the different versions of CAMNet architectures we compared for classification. BaseCNN is a basic Convolutional Neural Network with 6 convolutional layers followed by 3 fully-connected layers. BaseCNN2 is a deeper network with 10 convolutional layers which carries a similar amount of parameters as CAMNet3 in total (2.0m). MultiCNNX stands for X number of parallel BaseCNNs sharing the same input and the output (Averaging). In particular, we used 3 such BaseCNNs (MultiCNN3).

CAMNet architectures are designed such that an equivalent single path network would be a BaseCNN. We compared 3 versions of CAMNet by varying the number of parallel tensors in a layer (CAMNet2, CAMNet3, CAMNet4 where X in CAMNetX stands for number of parallel tensors in each layer). tinyCAMNet uses half the number of CNN filters in each layer, dropping two hidden fully-connected layers which has fewer parameters than BaseCNN.  Figure\ref{fig:camnet_classi} illustrates CAMNet3 architecture.

In addition, we build an SENet \cite{hu2017squeeze} equivalent to the BaseCNN, with the addition of gated weighting operations in each channel after convolutions as in the original paper. We also build a cross-stitch network (Cr-Stitch2) \cite{cross_stich} with two parallel paths, each being equal to a BaseCNN, with stitching operations wherever CAMNet2 has a routing layer instead. 

For image-to-image translation, we compare a shallow and a deep version of CAMNet. The shallow version has only routing layers, but no forward layers. In the deep version, a forward layer of similar size exists after each routing layer. We compare these versions of CAMNet with UNet architecture \cite{ronneberger2015unet} variant used in pix2pix implementation \cite{isola2017image}. This Unet variant carries 13 million parameters which is similar to the deep version of CAMNet. The shallow version has only 6 million parameters in contrast. We consider CAMNet versions with two parallel tensors in each layer in this evaluation.


\begin{table*}[t]
	\caption{Classification errors in individual and joint datasets: Row 5 shows the joint performance of all MNIST datasets.}
	\vspace{-0.2in}
	\label{tab:classi}
	\begin{center}
		\renewcommand{\tabcolsep}{0.8mm}
		\begin{tabular}[width=\columnwidth]{@{}lcccccccccr@{}}
			\toprule
			
			Network       & BaseCNN    	&\multicolumn{2}{c}{MultiCNN3}  & CAMNet2  &\multicolumn{2}{c}{CAMNet3}  &tinyCAMNet3 & CAMNet4  &SENet  &Cr-Stitch2\\
			Params(M) &0.5 &\multicolumn{2}{c}{1.5} &1.2 &\multicolumn{2}{c}{2.0} &0.47 &3.0 &0.5 &1 \\
			\midrule
			&       &\color{red}{No Aug.}&   &      &\color{red}{No Aug.}  & & & & & \\
			MNIST     	&0.3 	&\color{red}{0.48} &0.3   &0.25  &\color{red}{0.53} &\textbf{0.22} &0.32 &0.26   &0.25  &0.33\\
			Fashion   	    &6.37	&\color{red}{6.53} &6.02  &5.8   &\color{red}{7.0} &\textbf{5.66}    &5.98 &5.69   &5.92  &5.88\\
			KMNIST      	    &1.38	&\color{red}{2.73} &1.14  &1.07  &\color{red}{2.52} &\textbf{0.95} &1.26 &1.06   &1.15  &1.16\\
			NotMNIST         &2.37   &\color{red}{3.1} &2.54   &2.39  &\color{red}{3.38} &2.18              &2.64 &\textbf{2.15}  &2.47  &2.47\\
			\midrule
			Joint       &2.93   &\color{red}{4.17} &3.03  &2.66  &\color{red}{3.84} &2.57 &2.82 & \textbf{2.52} &2.88  &2.98\\
			\midrule
			CIFAR10      &8.3 & &9.71 &7.59 & &\textbf{7.02}  &8.88 &7.06  &8.71  &8.22\\
			SVHN         &3.86 & &3.23 &3.54 & &3.28  &3.38 &\textbf{3.20}  &3.42  &3.45\\
			\bottomrule
		\end{tabular}
	\end{center}
	\vspace{-0.1in}
\end{table*}

\subsection{Individual and Joint Classification}
\label{ss:ind_data}
\vspace{-0.05in}
We evaluate our model with the  MNIST dataset where we record an error of \textbf{0.22\%} with CAMNet3 having 2.0 million parameters. We train with images which are augmented by pixel shift, rotation and scaling with no padding. Wan \etal\cite{dropconnet} achieve an error of 0.21\% with data augmentation and Sobour \etal\cite{sabour2017dynamic}, an error of 0.25\% error with pixel shift and reconstruction loss. 

Table \ref{tab:classi} further shows the error with other datasets where CAMNet outperforms other architectures. CAMNet surpasses the equivalent SENet because CAMNet can benefit from multiple parallel paths for a certain input, in contrast to a single path instance in SENet. CAMNet2 also surpasses the performance of Cross-Stitch Network (Cr-Stitch2), showing that data-dependent gating is superior than independently learned stitching, specially for data with varying context. 

In addition we jointly train our network with the aforementioned MNIST classification datasets (Table \ref{tab:classi}, $5^{th}$ row), where CAMNet4 performs the best in combination, achieving an overall error of \textbf{2.52\%}. We also compare the joint performance with \textbf{Rebuffi \etal} \cite{residual-adapters} (1.4M), which consists separate task-specific softmax layers at the output for each dataset. Rebuffi \etal shows joint error of \textbf{2.7\%} which is similar to CAMNet2, but surpassed by CAMNet3 and CAMNet4.

One important observation is tinyCAMNet3, managing to surpass MultiCNN3 in joint training even though its performance in individual datasets is similar (Table \ref{tab:classi}). This effect proves that CAMNets are particularly suitable for learning from multiple domains. We repeat individual and joint training \textbf{without augmentation} for CAMNet3 and MultiCNN3. Here, CAMNet3 shows the above observation profoundly (Table \ref{tab:classi} No Aug.* columns). This effect is being dampened when CAMNets are trained with data augmentation, showing that CAMNets capture the variation introduced by augmentation better than baselines to improve performance even in individual datasets.


\subsection{Lifelong Learning}
\label{ss:seq_mul}
\vspace{-0.05in}

\begin{figure}[t]
	\begin{center}
		\begin{subfigure}[b]{0.32\linewidth}
			\includegraphics[width=\linewidth]{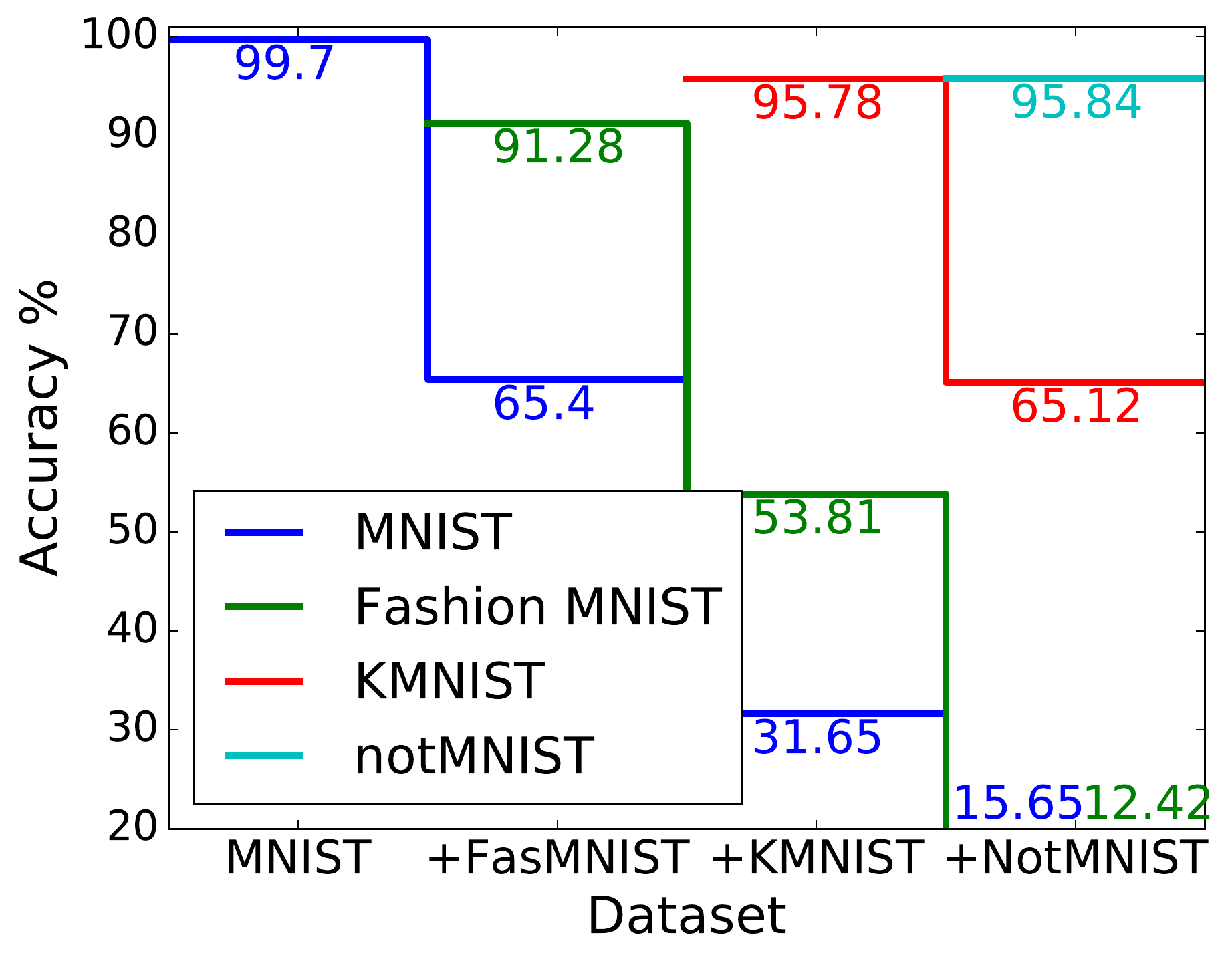} 
			\caption{BaseCNN}
		\end{subfigure}
		\begin{subfigure}[b]{0.32\linewidth}
			\includegraphics[width=\linewidth]{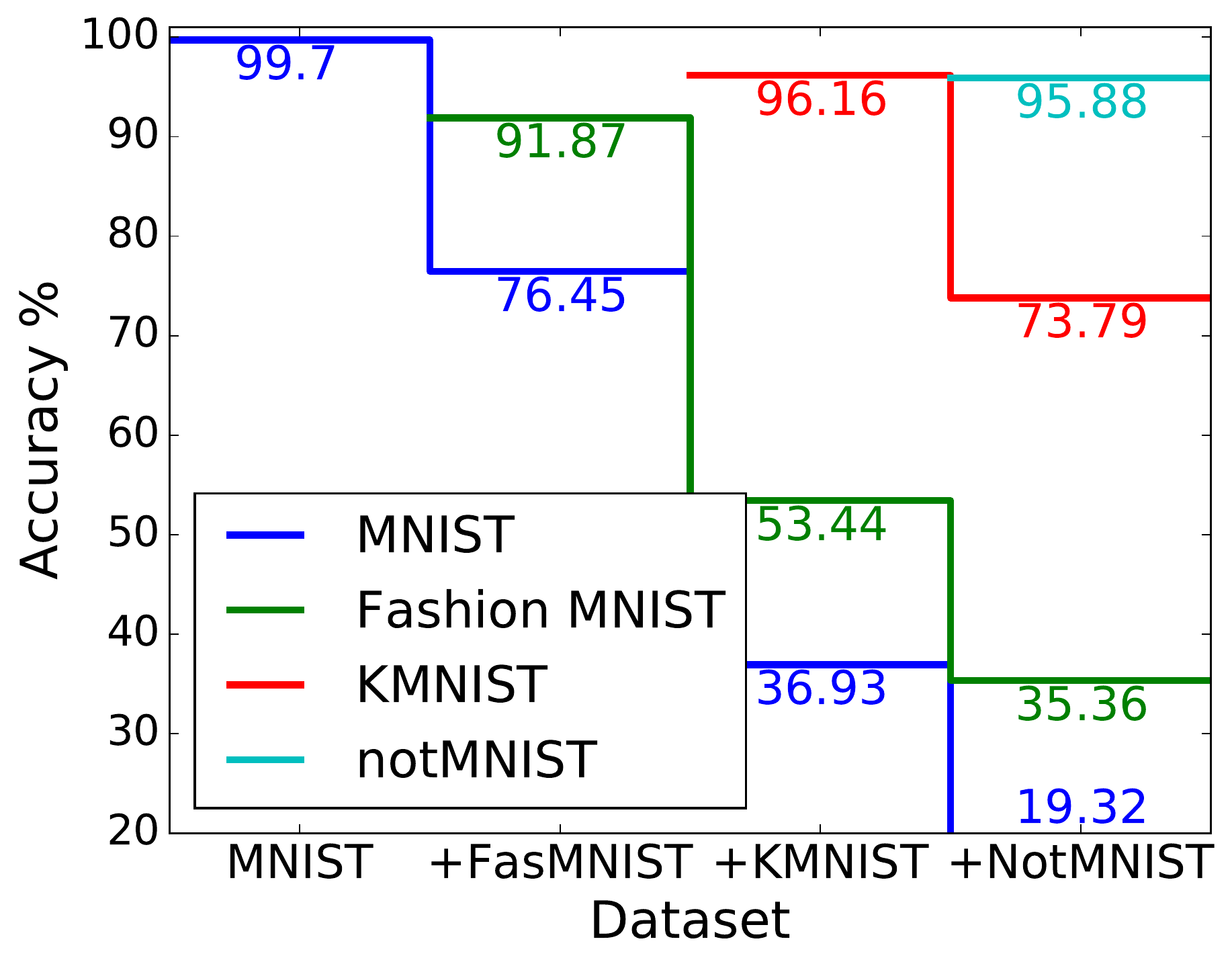}
			\caption{MultiCNN3}
		\end{subfigure}
		\begin{subfigure}[b]{0.32\linewidth}
			\includegraphics[width=\linewidth]{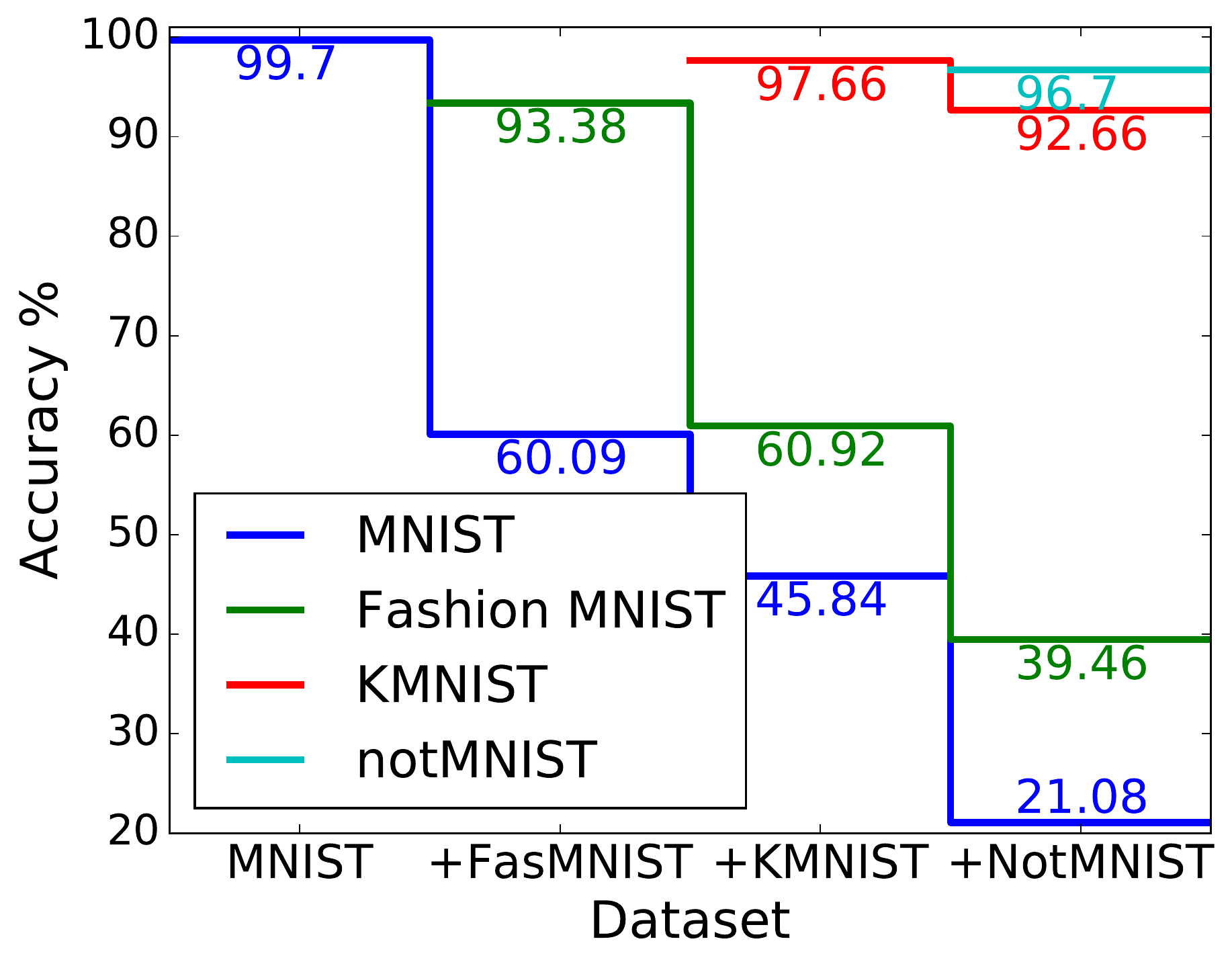}
			\caption{BaseCNN2}
		\end{subfigure}
		\begin{subfigure}[b]{0.32\linewidth}
			\includegraphics[width=\linewidth]{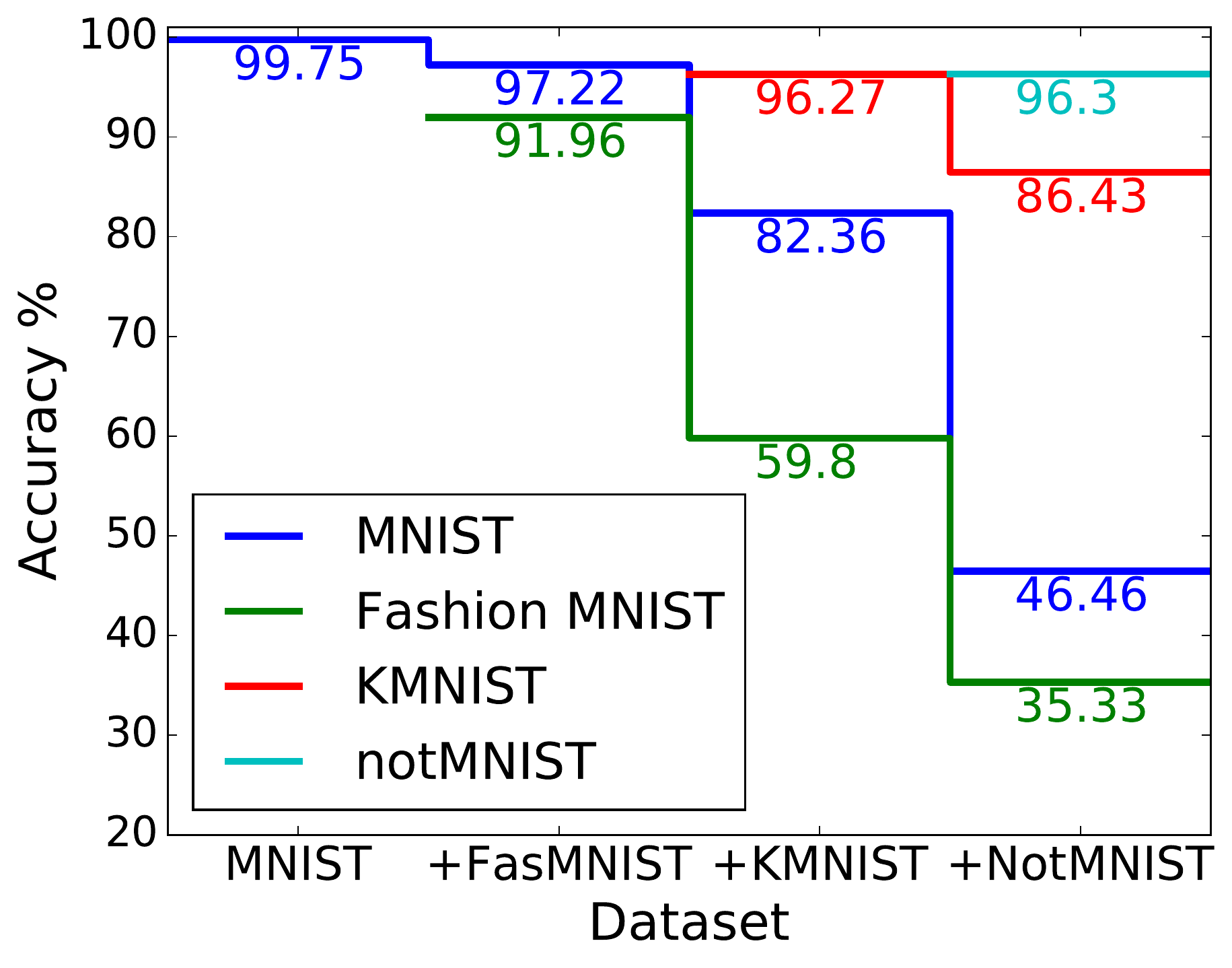}
			\caption{CAMNet2}
		\end{subfigure}
		\begin{subfigure}[b]{0.32\linewidth}
			\includegraphics[width=\linewidth]{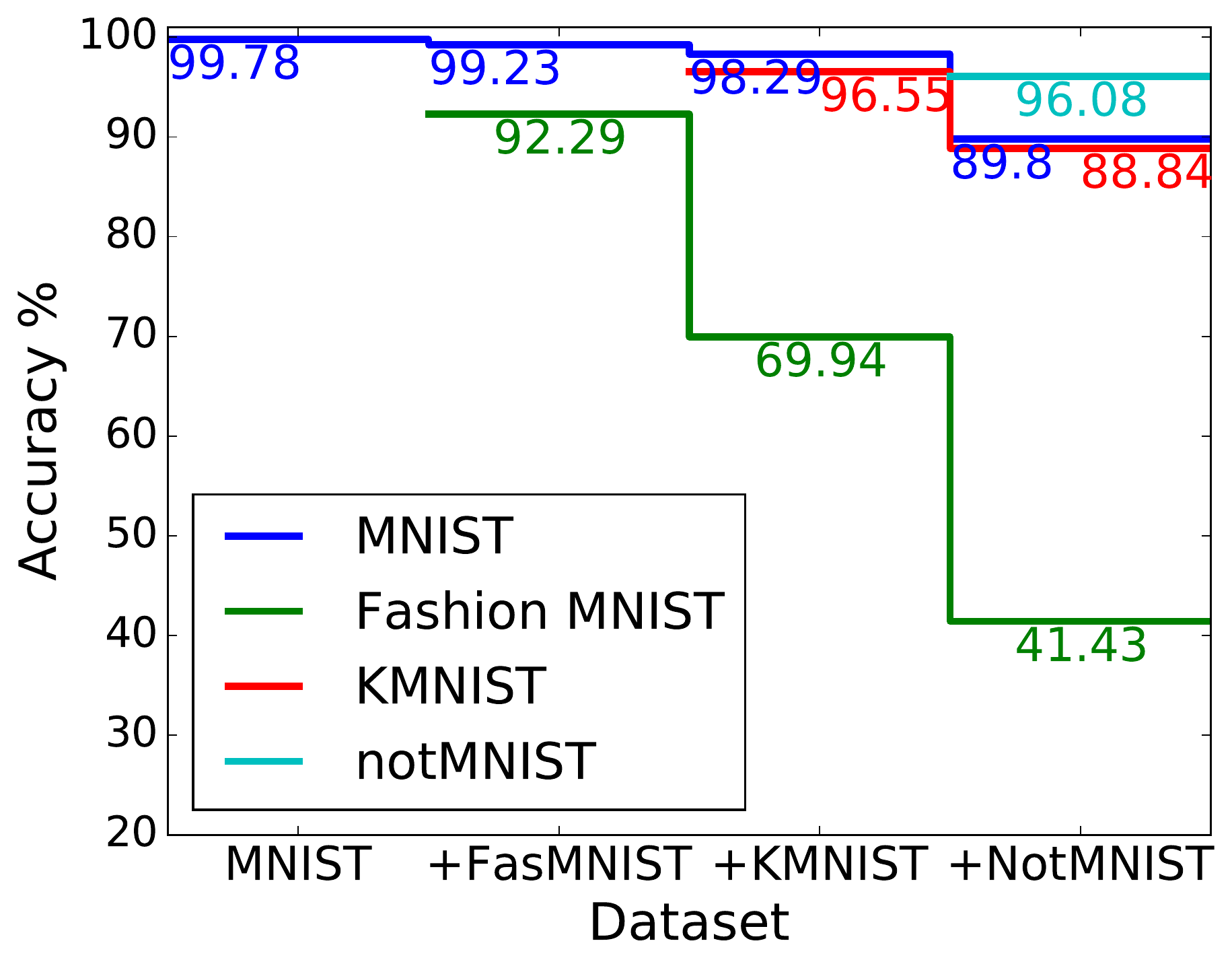} 
			\caption{CAMNet3}
		\end{subfigure}
		\begin{subfigure}[b]{0.32\linewidth}
			\includegraphics[width=\linewidth]{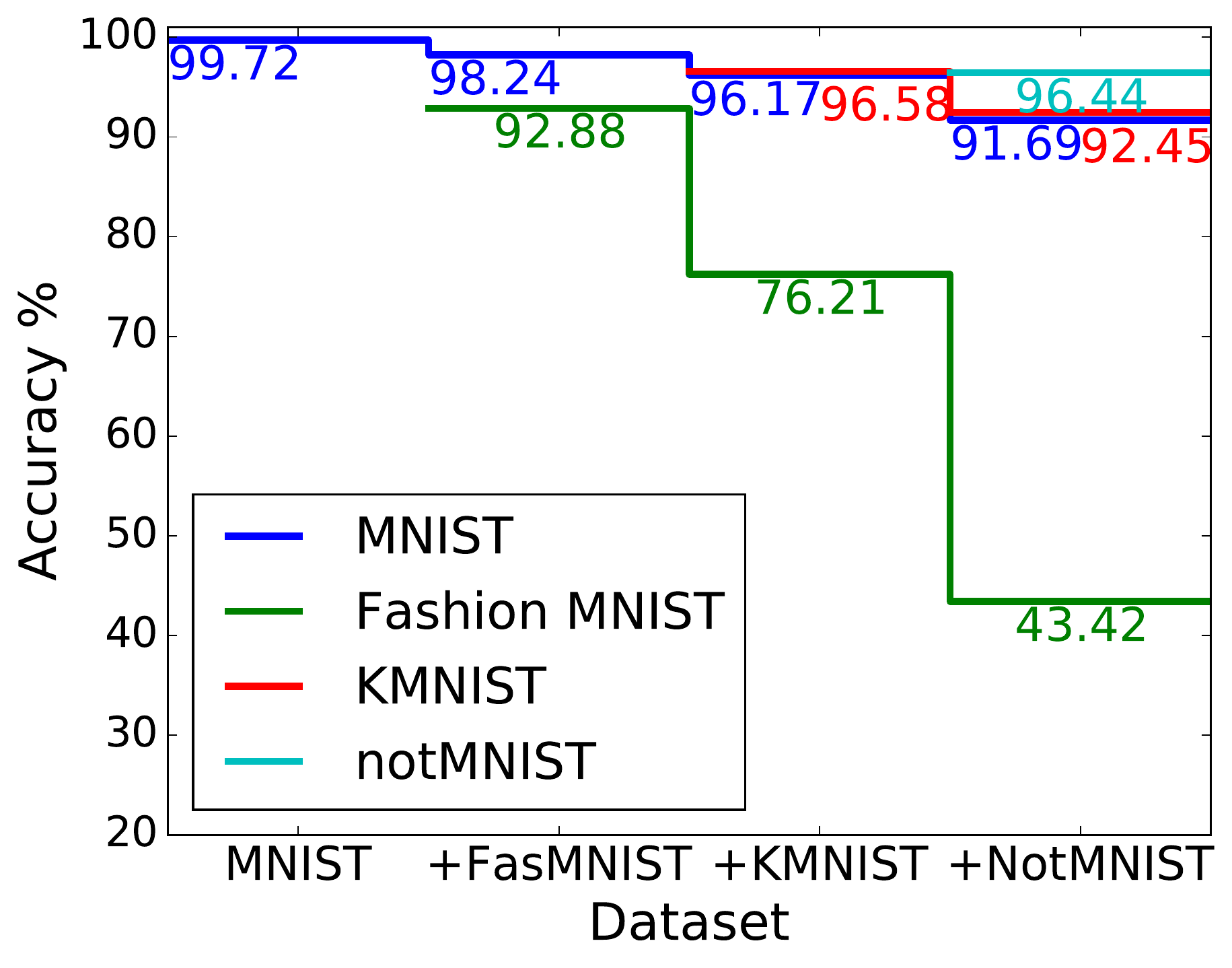}
			\caption{CAMNet4}
		\end{subfigure}
	\end{center}
	\vspace{-0.2in}
	\caption{Accuracy change when trained on multiple subsequent datasets: CAMNet architectures preserve the learned features better, in comparison with BaseCNN, MultiCNN, and even a deeper CNN (BaseCNN2). The lifelong leraning results are averaged over 3 trials}
	\label{fig:lwf_plots}
	\vspace{-0.1in}
\end{figure}

\begin{figure}[ht]
	\begin{center}
		\begin{subfigure}[b]{0.32\columnwidth}
			\includegraphics[width=\linewidth]{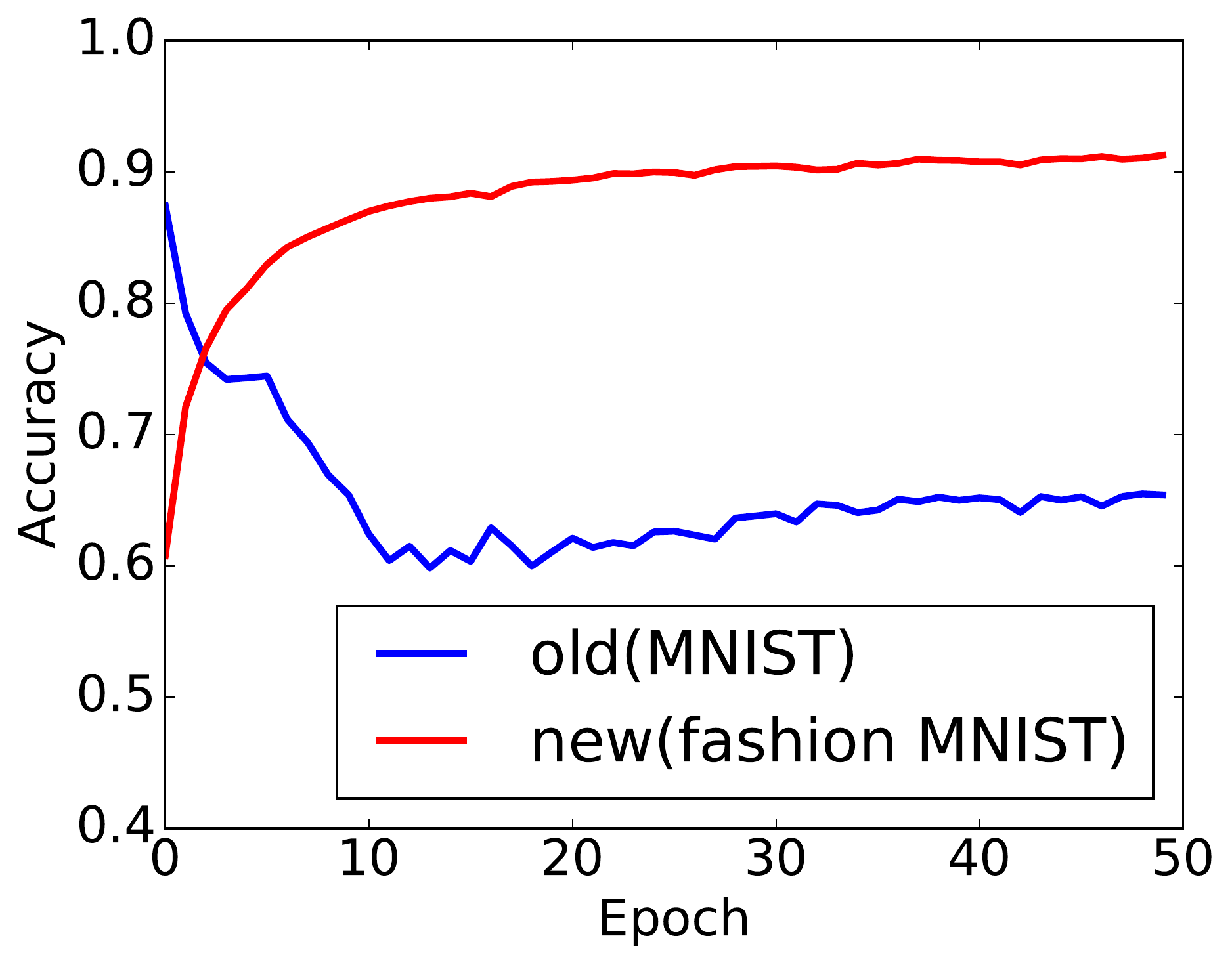}
			\caption{BaseCNN}
		\end{subfigure}
		\begin{subfigure}[b]{0.32\columnwidth}
			\includegraphics[width=\linewidth]{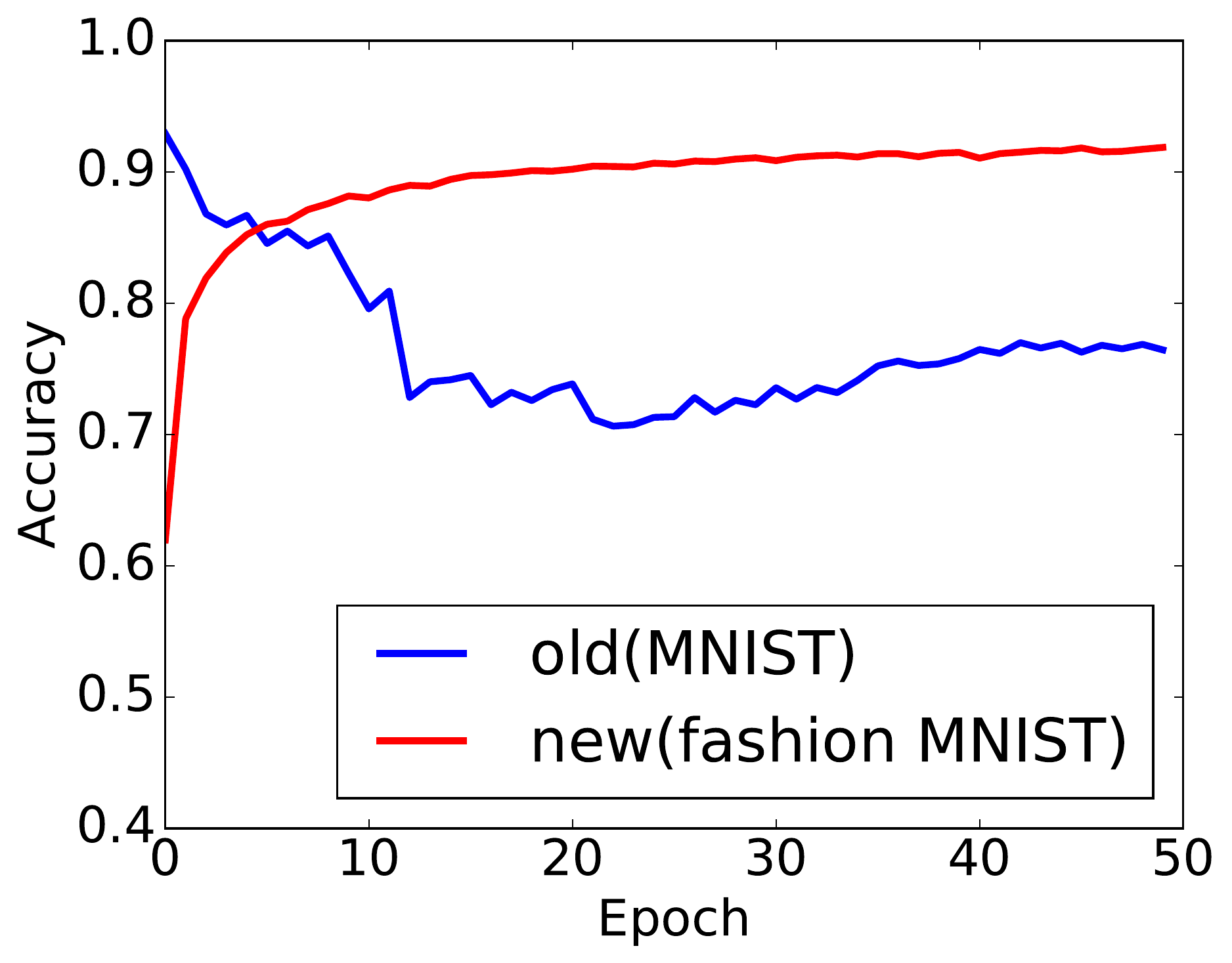}
			\caption{MultiCNN3}
		\end{subfigure}
		\begin{subfigure}[b]{0.32\columnwidth}
			\includegraphics[width=\linewidth]{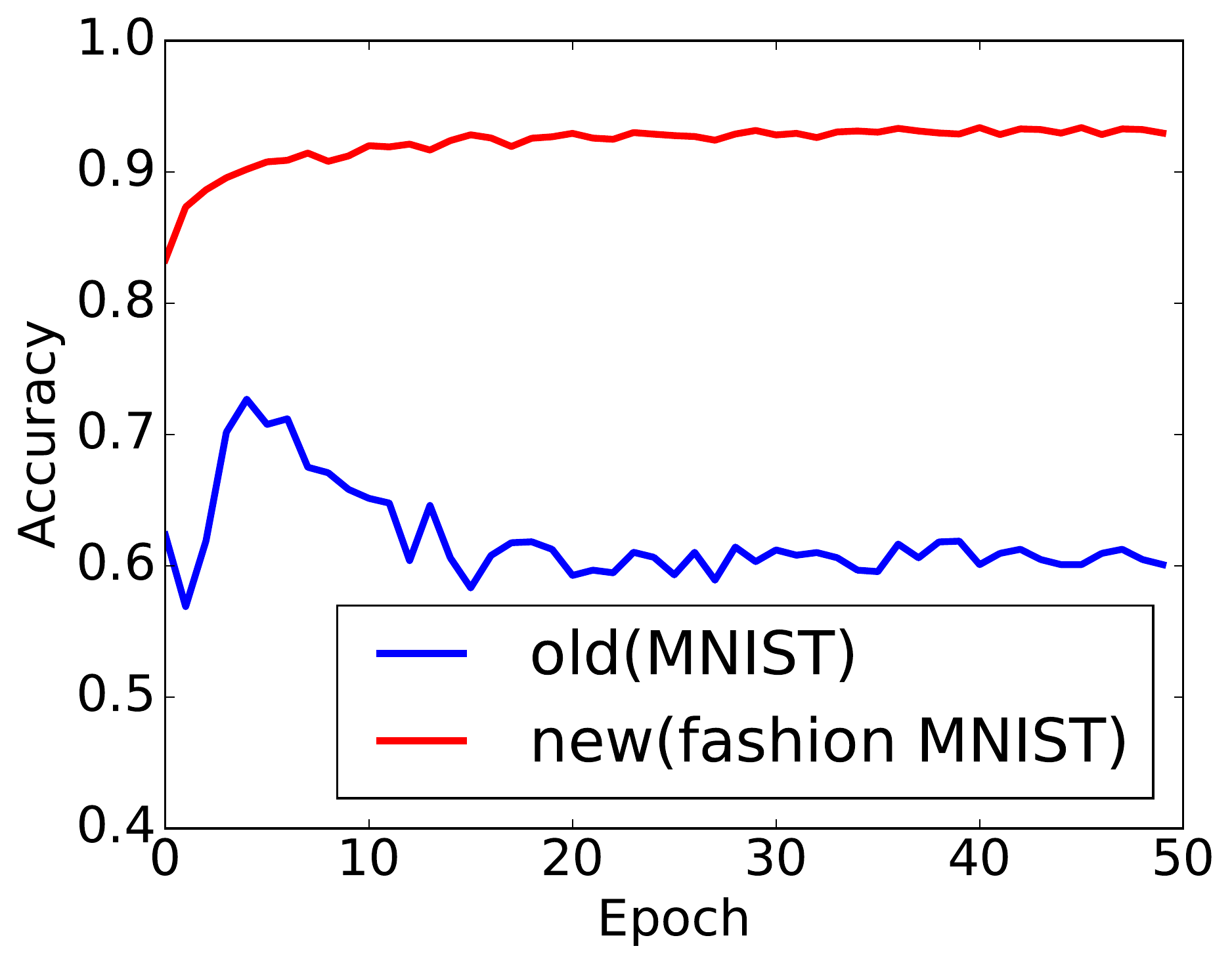}
			\caption{BaseCNN2}
		\end{subfigure}
		\begin{subfigure}[b]{0.32\columnwidth}
			\includegraphics[width=\linewidth]{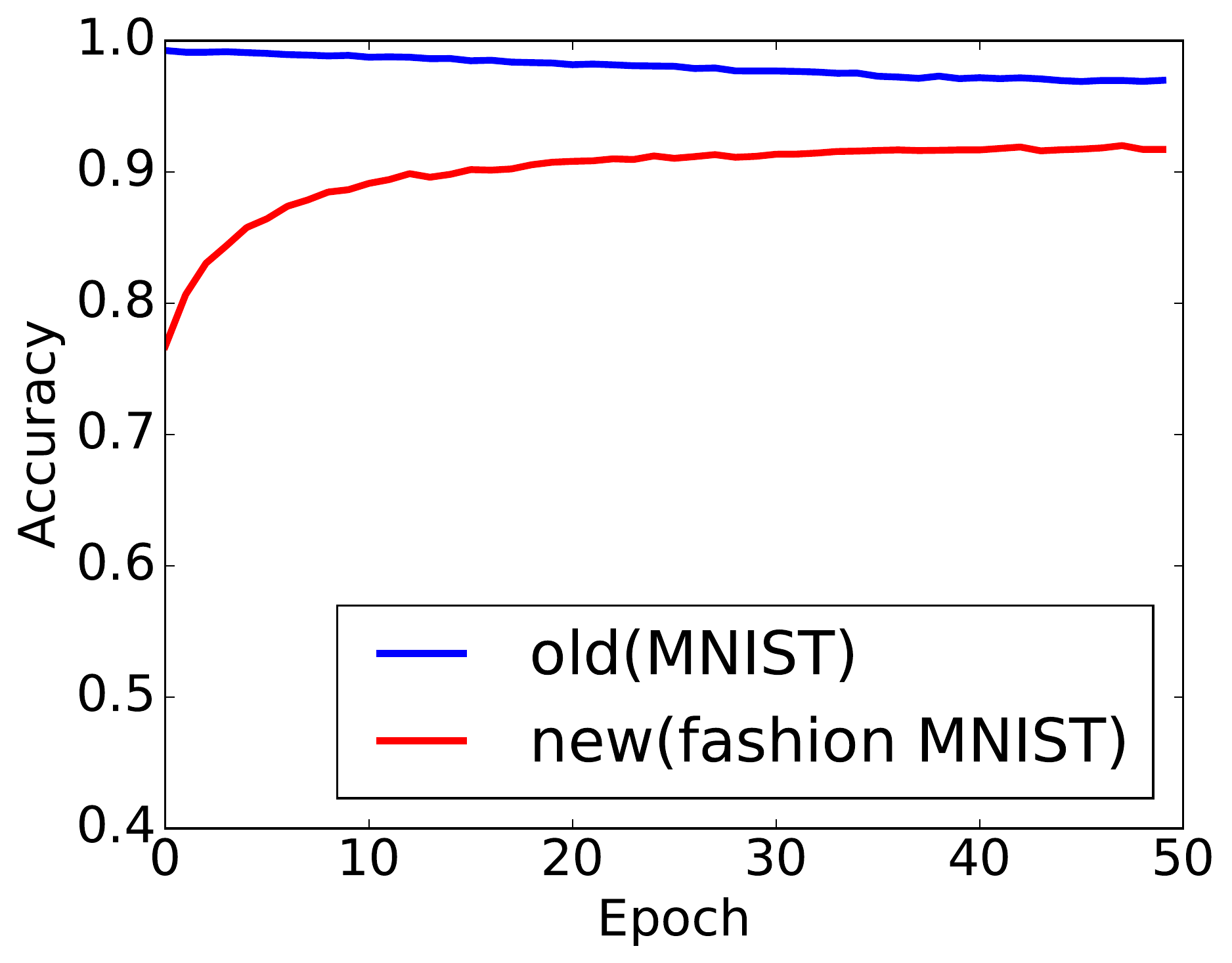}
			\caption{CAMNet2}
		\end{subfigure}
		\begin{subfigure}[b]{0.32\columnwidth}
			\includegraphics[width=\linewidth]{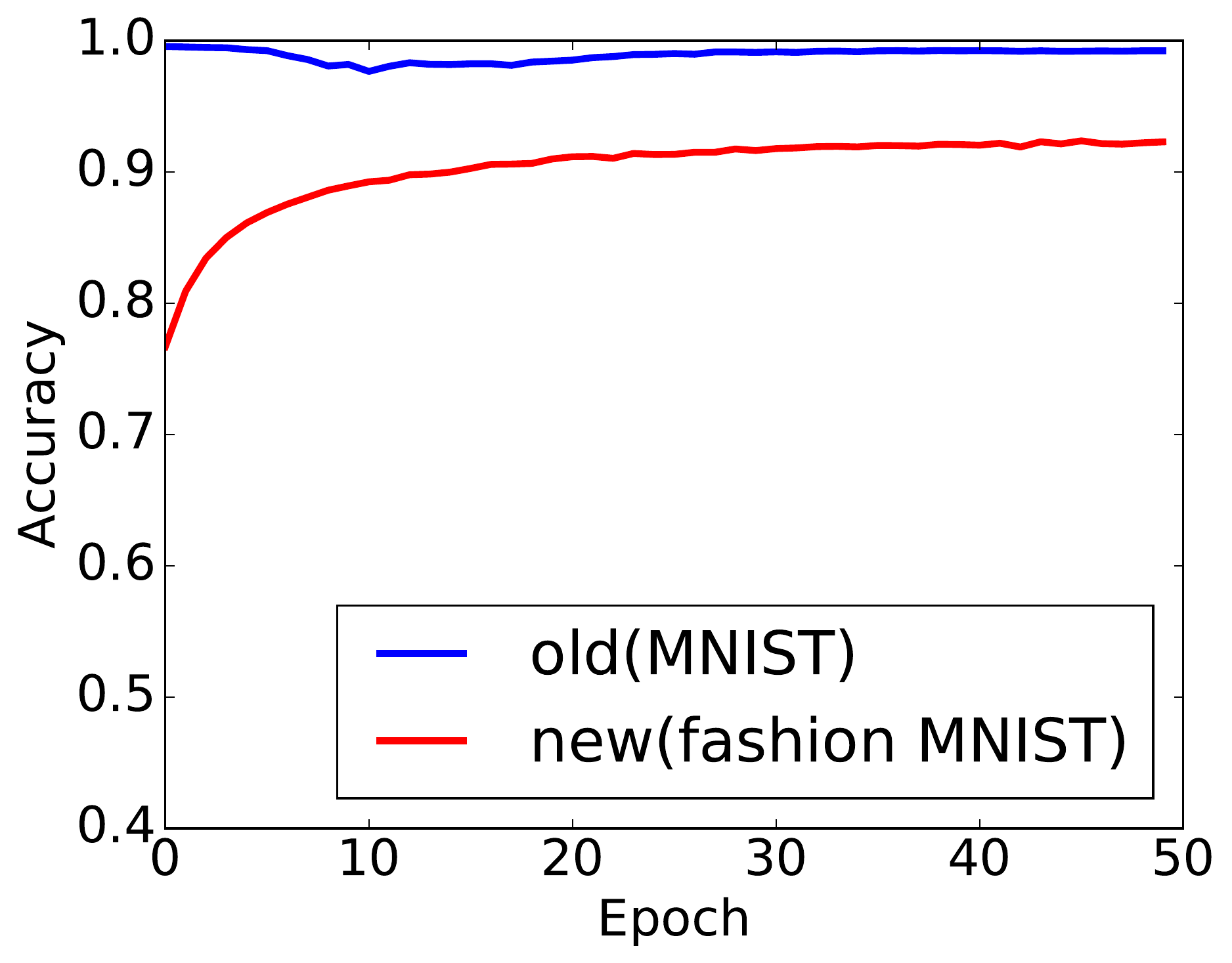}
			\caption{CAMNet3}
		\end{subfigure}
		\begin{subfigure}[b]{0.32\columnwidth}
			\includegraphics[width=\linewidth]{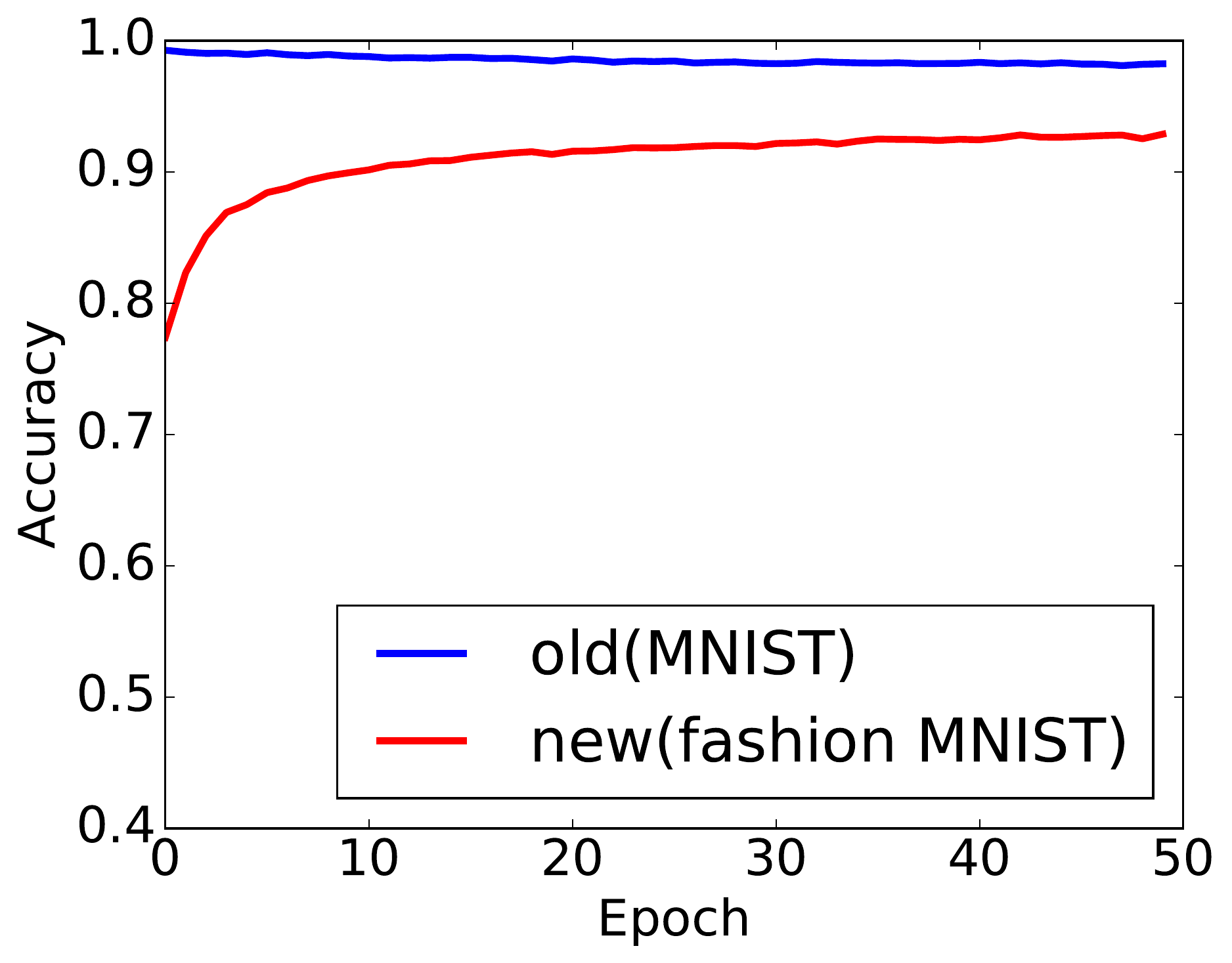}
			\caption{CAMNet4}
		\end{subfigure}
	\end{center}
	\vspace{-0.2in}
	\caption{Accuracy change when trained on a subsequent dataset. CAMNet architectures show a smoother adaptation to the second dataset while preserving the accuracy on the first.}
	\label{fig:learning_curve}
	\vspace{-0.1in}
\end{figure}

We adopt the Learning without Forgetting (LwF) technique (see Section \ref{ss:sa_training}) to train our model with the four classification datasets considered, sequentially. In this experiment, we feed the datasets in the order of MNIST, Fashion MNIST, KMNIST, and NotMNIST. Figure \ref{fig:lwf_plots} illustrates the change of the test accuracy on previous datasets after trained on a new dataset. CAMNet3 simply outperforms its equivalent single path CNN (BaseCNN), equivalent multipath CNN (MultiCNN3) and deeper BaseCNN2, whereas CAMNet4 shows the best results. In addition, CAMNet architectures shows a smooth adaptation to the Fashion MNIST dataset once trained on the MNIST dataset. (See the curves in Figure \ref{fig:learning_curve})

\subsection{Image-to-image Translation}
\label{ss:im_to_im}
\vspace{-0.05in}
We develop the combined dataset for joint training with all the pixel labelling datasets (see Section \ref{ss:ex_datasets} for datasets) by stacking each dataset, 5\% at a time. The models are trained as an image-to-image translation with mean squared error (L2 loss) on this joint dataset, despite some datasets being segmentation datasets. All the data from different datasets are re-scaled to a common resolution of $256\times256$. We further evaluate the model with satellite-to-aerial (Drones) dataset and drone image-to-road topology (Maps) dataset, individually.

\begin{table}[ht]
	\caption{Test loss comparison of image-to-image translation with satellite-to-aerial (Drones) dataset and drone image-to-road topology (Maps) dataset individually, and the combined dataset (Drones, Maps, Facades, Cityscapes, KITTI). DeepCAMNet reduces the error significantly compared to UNet-pix2pix.}
	\label{tab:imtoim}
	\vspace{-0.2in}
	\begin{center}
		\begin{tabular}[width=\columnwidth]{@{}lccr@{}}
			\toprule
			Dataset 			& Drones & Maps	& Combined \\
			\midrule
			UNet-pix2pix     	&0.0785 &0.0133 &0.0798	\\
			\midrule
			CAMNet   			&-0.51\%    &-6.02\%	&-7.27\%	\\
			DeepCAMNet       	&\textbf{-14.65}\% 		&\textbf{-18.8}\%   &\textbf{-14.16\%}	\\
			\bottomrule
		\end{tabular}
	\end{center}
	\vspace{-0.1in}
\end{table} 

\begin{figure}[htbp]
	\begin{center}
		\begin{subfigure}[b]{0.24\linewidth}
			\includegraphics[width=\linewidth]{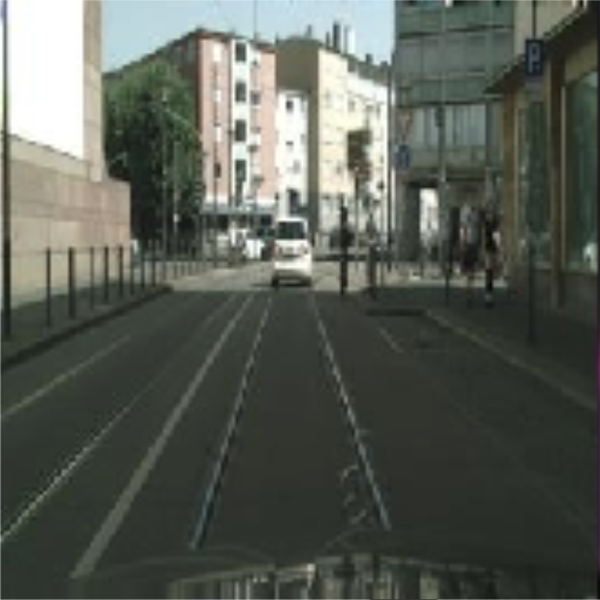} 
		\end{subfigure}
		\begin{subfigure}[b]{0.24\linewidth}
			\includegraphics[width=\linewidth]{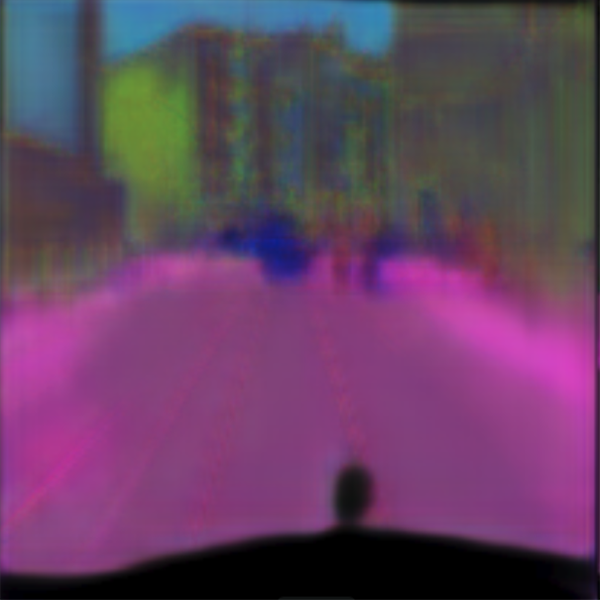}
		\end{subfigure}
		\begin{subfigure}[b]{0.24\linewidth}
			\includegraphics[width=\linewidth]{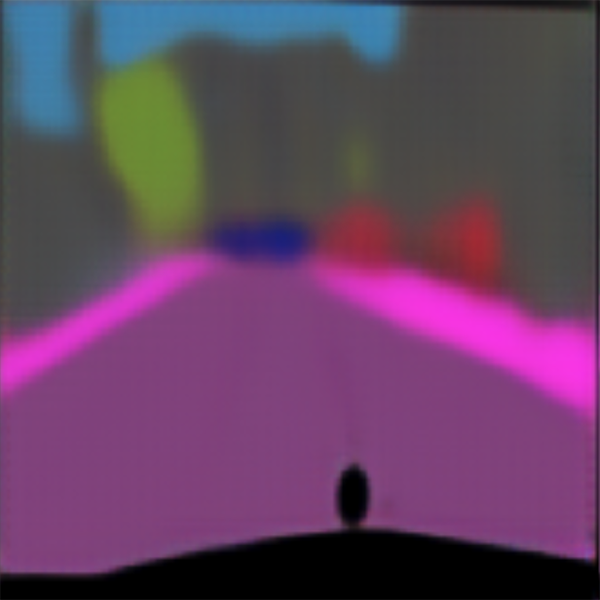}
		\end{subfigure}
		\begin{subfigure}[b]{0.24\linewidth}
			\includegraphics[width=\linewidth]{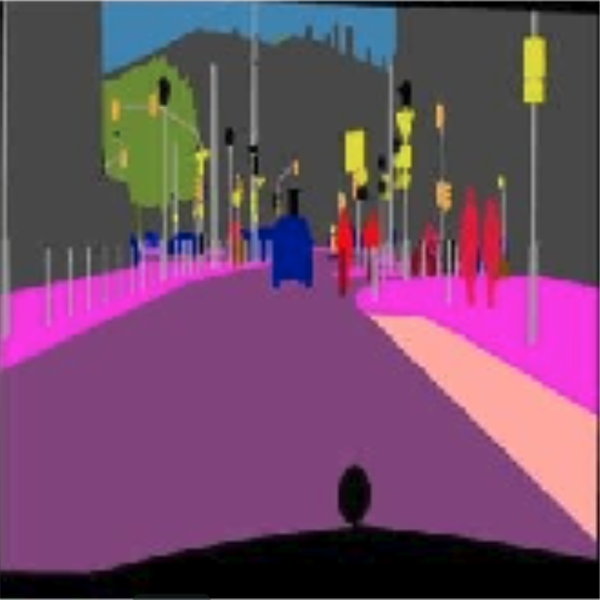}
		\end{subfigure}
		
		\begin{subfigure}[b]{0.24\linewidth}
			\includegraphics[width=\linewidth]{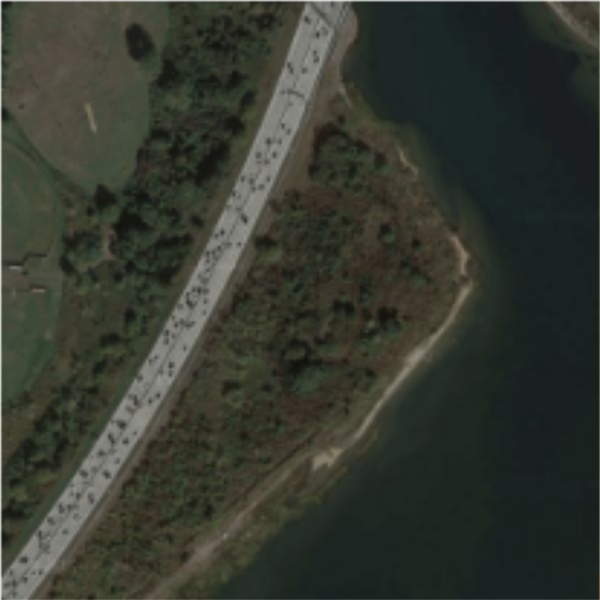} 
		\end{subfigure}
		\begin{subfigure}[b]{0.24\linewidth}
			\includegraphics[width=\linewidth]{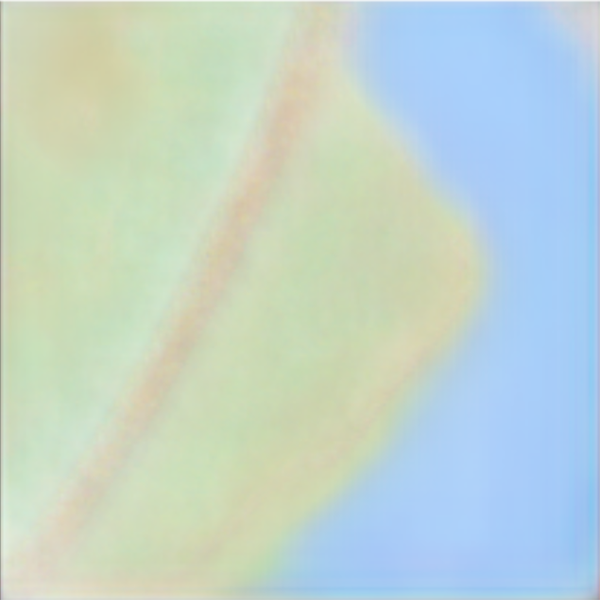}
		\end{subfigure}
		\begin{subfigure}[b]{0.24\linewidth}
			\includegraphics[width=\linewidth]{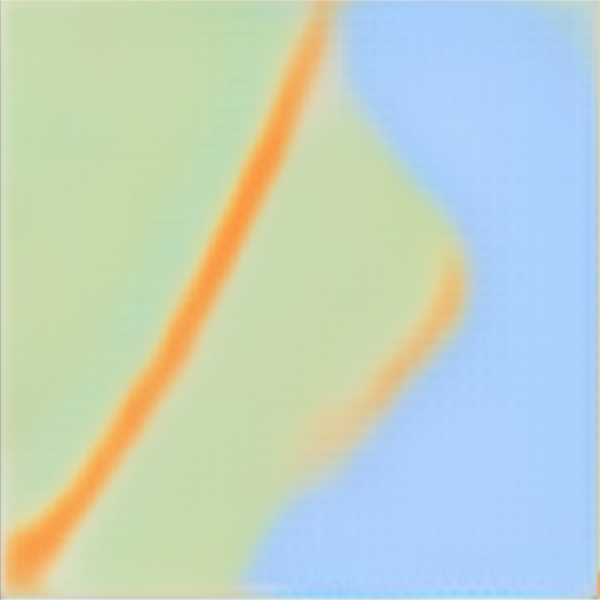}
		\end{subfigure}
		\begin{subfigure}[b]{0.24\linewidth}
			\includegraphics[width=\linewidth]{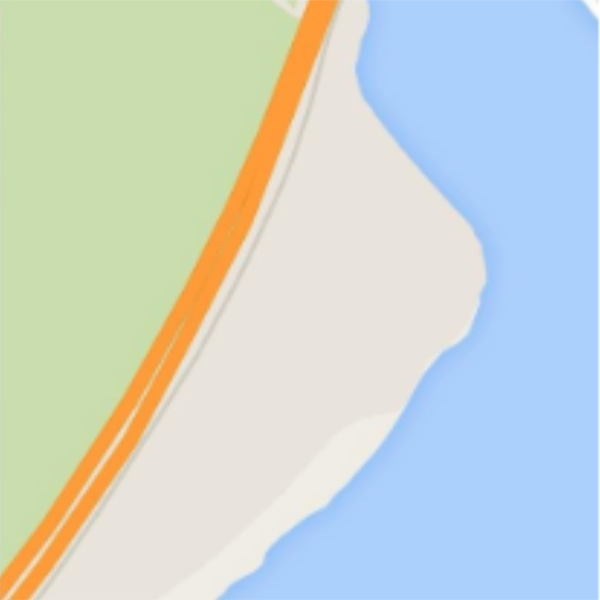}
		\end{subfigure}
		
		\begin{subfigure}[b]{0.24\linewidth}
			\includegraphics[width=\linewidth]{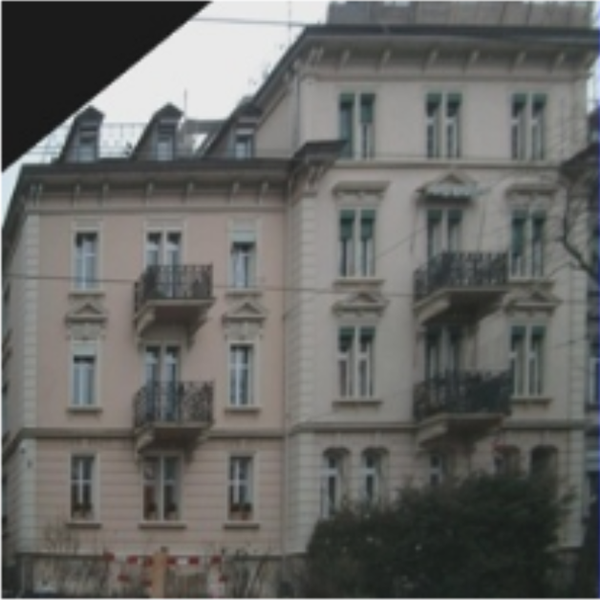} 
		\end{subfigure}
		\begin{subfigure}[b]{0.24\linewidth}
			\includegraphics[width=\linewidth]{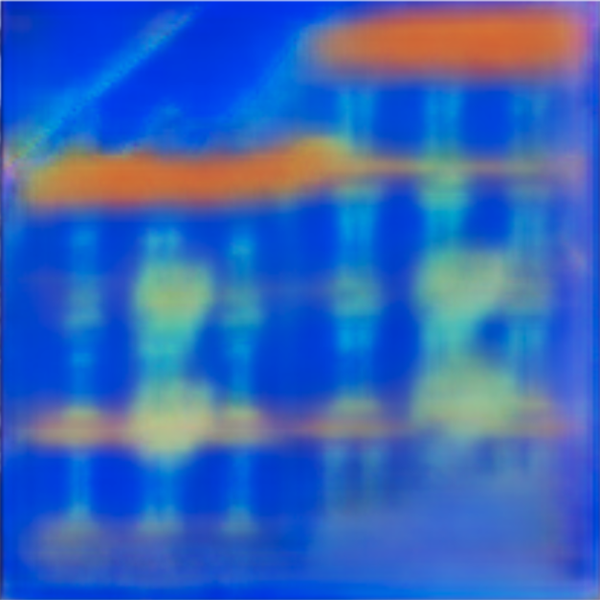}
		\end{subfigure}
		\begin{subfigure}[b]{0.24\linewidth}
			\includegraphics[width=\linewidth]{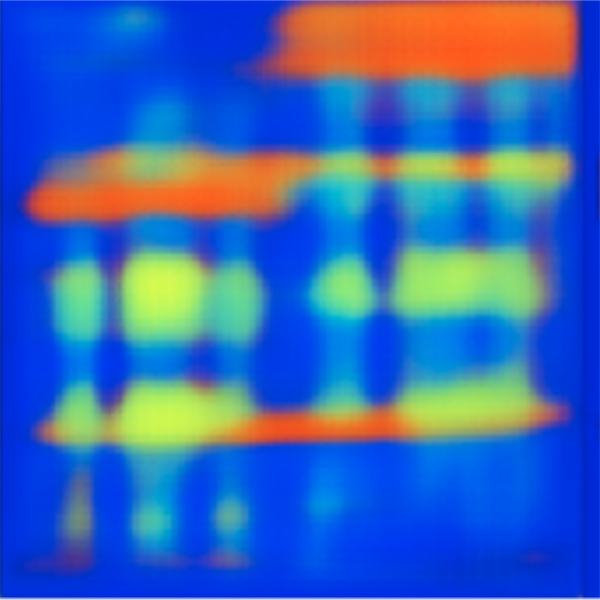}
		\end{subfigure}
		\begin{subfigure}[b]{0.24\linewidth}
			\includegraphics[width=\linewidth]{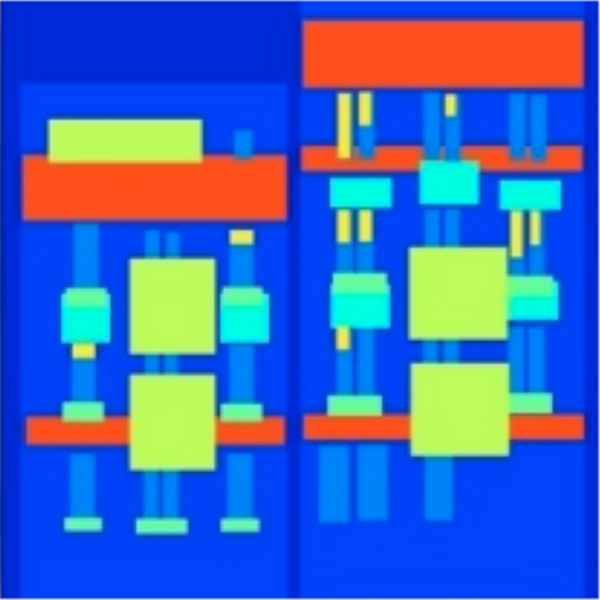}
		\end{subfigure}
		
		\begin{subfigure}[b]{0.24\linewidth}
			\includegraphics[width=\linewidth]{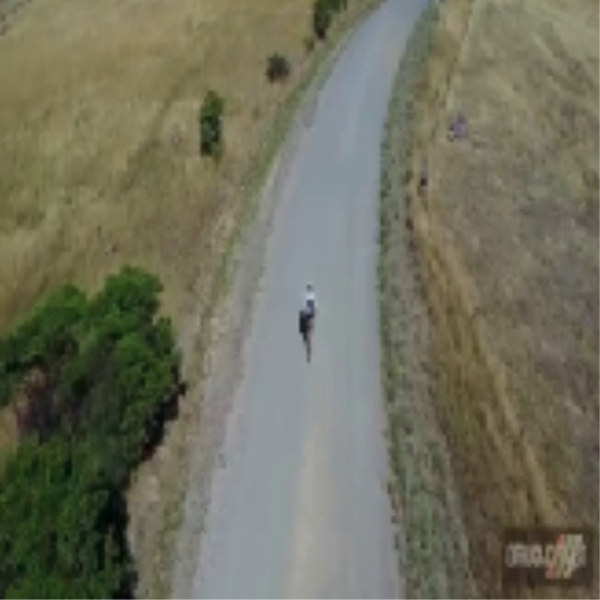} 
		\end{subfigure}
		\begin{subfigure}[b]{0.24\linewidth}
			\includegraphics[width=\linewidth]{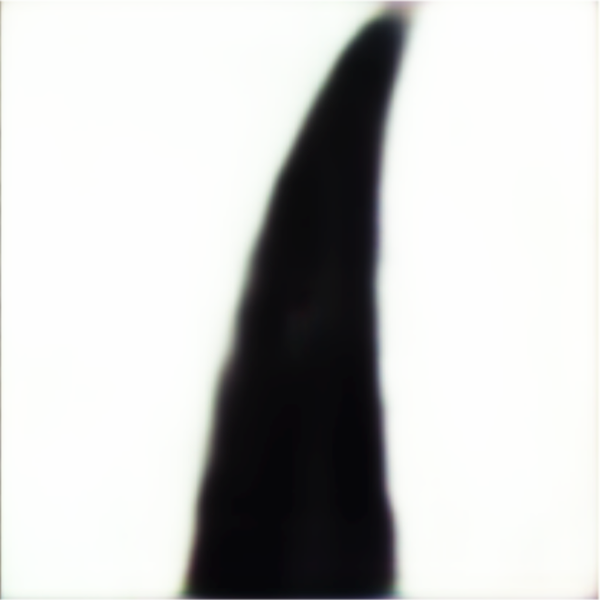}
		\end{subfigure}
		\begin{subfigure}[b]{0.24\linewidth}
			\includegraphics[width=\linewidth]{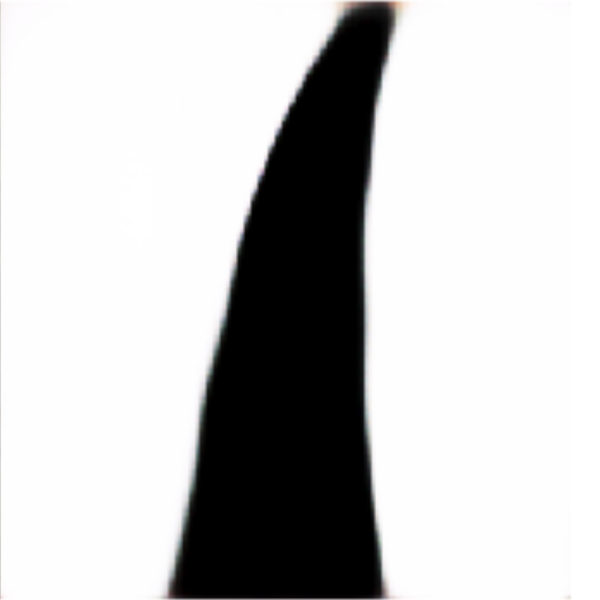}
		\end{subfigure}
		\begin{subfigure}[b]{0.24\linewidth}
			\includegraphics[width=\linewidth]{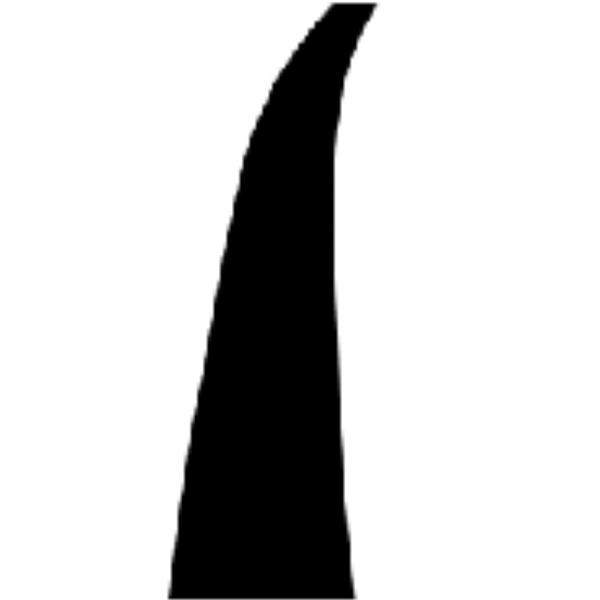}
		\end{subfigure}
		
		\begin{subfigure}[b]{0.24\linewidth}
			\includegraphics[width=\linewidth]{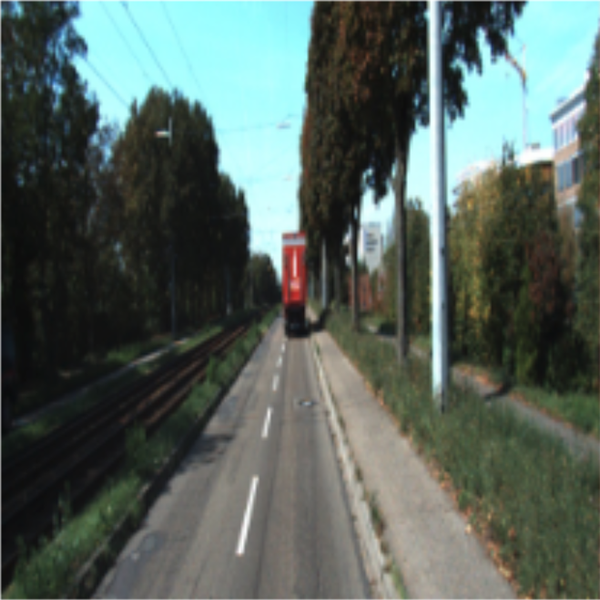}
			\caption{Input} 
		\end{subfigure}
		\begin{subfigure}[b]{0.24\linewidth}
			\includegraphics[width=\linewidth]{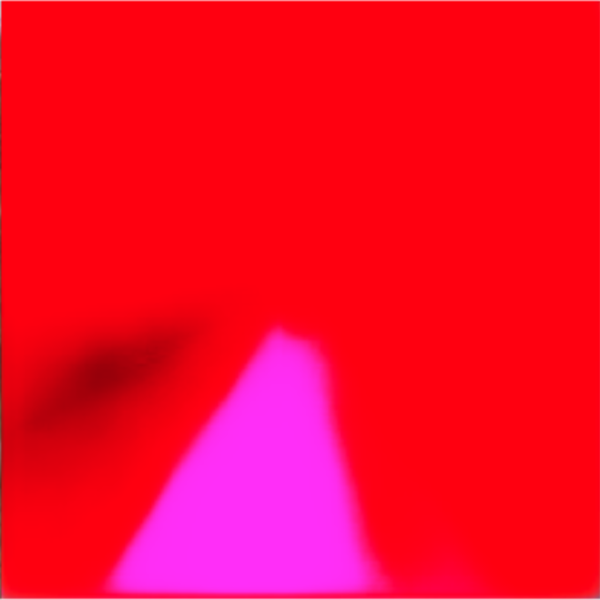}
			\caption{UNet}
		\end{subfigure}
		\begin{subfigure}[b]{0.24\linewidth}
			\includegraphics[width=\linewidth]{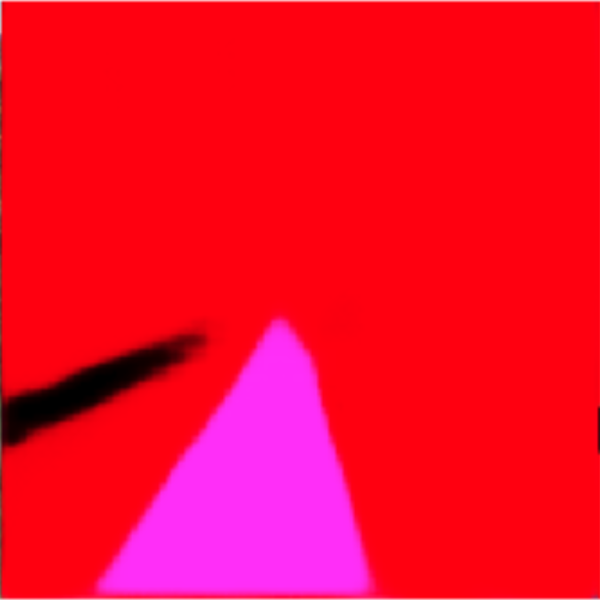}
			\caption{CAMNet}
		\end{subfigure}
		\begin{subfigure}[b]{0.24\linewidth}
			\includegraphics[width=\linewidth]{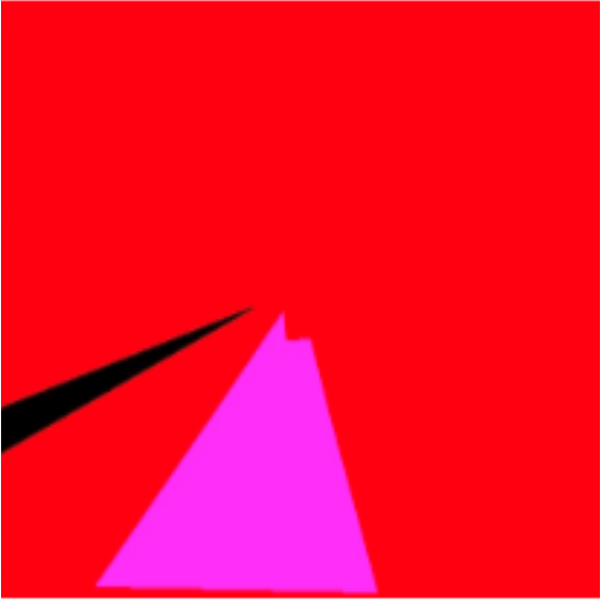}
			\caption{Target}
		\end{subfigure}
		
	\end{center}
	\vspace{-0.2in}
	\caption{Outputs of Unet-pix2pix and deep CAMNet after joint training in the combined dataset. Deep CAMNet shows better outputs, closer to the target.}
	\label{fig:imtoim}
	\vspace{-0.1in}
\end{figure}

Table \ref{tab:imtoim} illustrates the versions of CAMNet (Sec. \ref{ss:compared_archi}) outperforming the UNet architecture in individual and combined datasets. We report the changes in test loss compared to UNet. Figure \ref{fig:imtoim} shows the visual comparison of results between UNet-pix2pix and deep CAMNet trained on the combined dataset.

\begin{figure}[ht]
	\begin{center}
		\begin{subfigure}[b]{\columnwidth}
			\includegraphics[width=\linewidth]{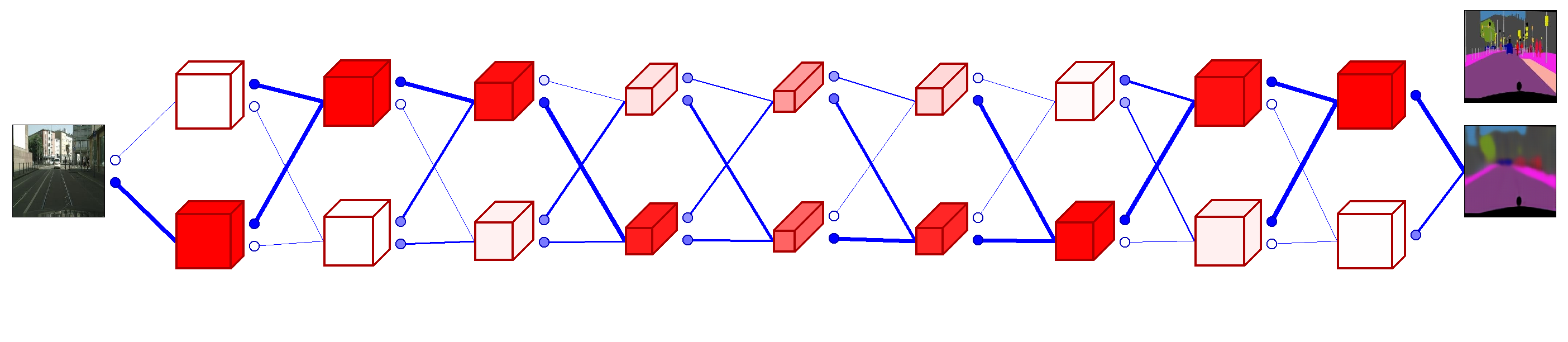}
		\end{subfigure}
		\begin{subfigure}[b]{\columnwidth}
			\includegraphics[width=\linewidth]{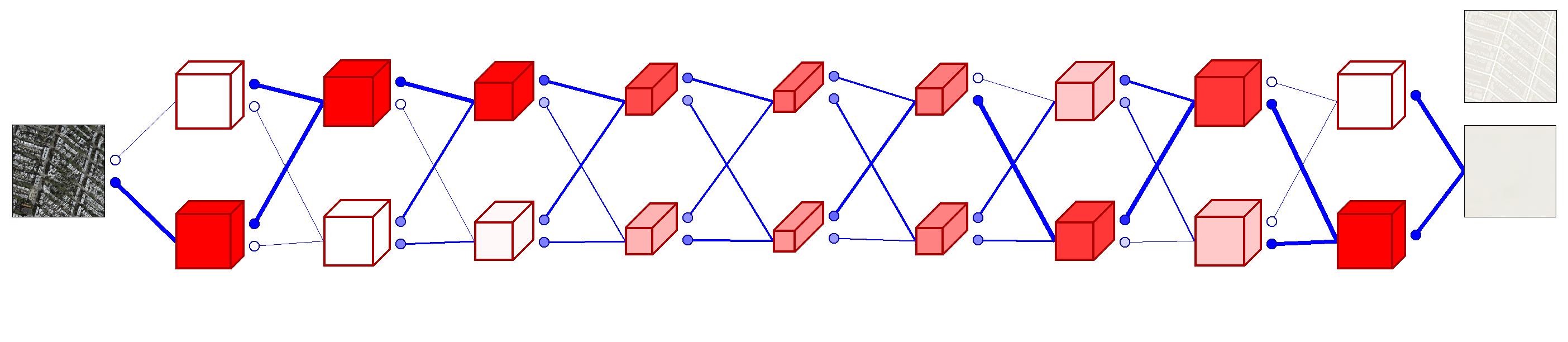}
		\end{subfigure}
		\begin{subfigure}[b]{\columnwidth}
			\includegraphics[width=\linewidth]{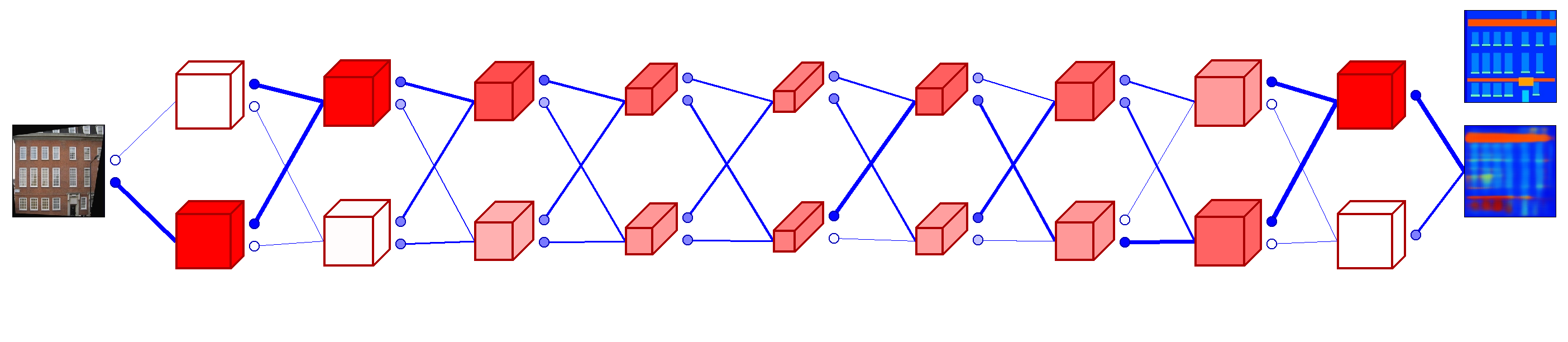}
		\end{subfigure}
		\begin{subfigure}[b]{\columnwidth}
			\includegraphics[width=\linewidth]{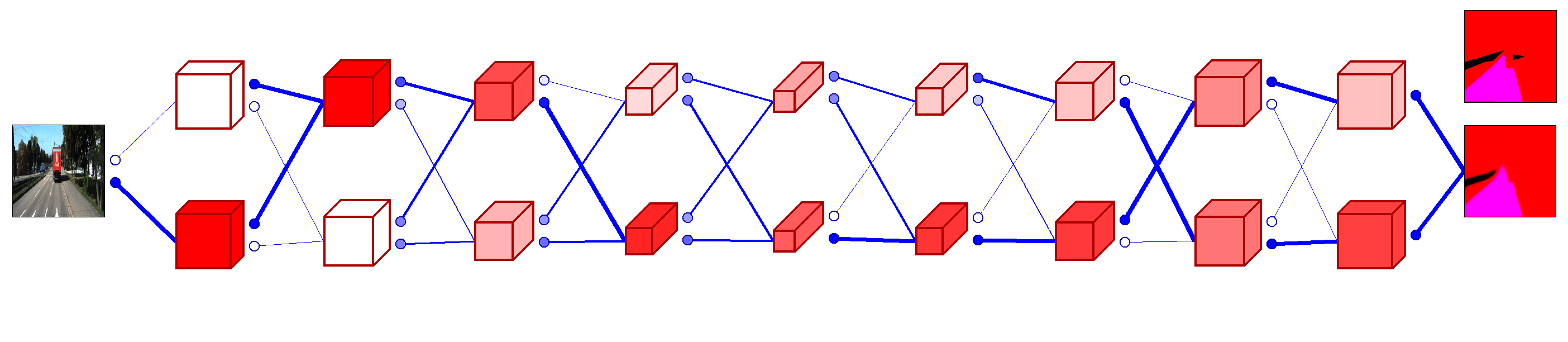}
		\end{subfigure}
		\begin{subfigure}[b]{\columnwidth}
			\includegraphics[width=\linewidth]{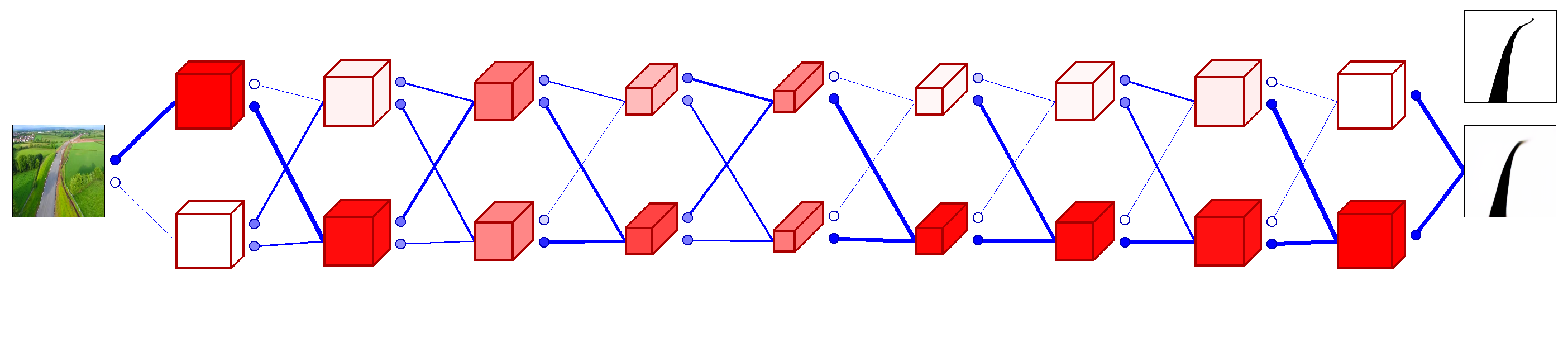}
		\end{subfigure}
	\end{center}
	\vspace{-0.2in}
	\caption{Route Visualization in image-to-image translation. Different path combinations considered for each dataset after joint training. The dark blue connections and dark red tensors show the paths taken with frequently.}
	\label{fig:imtoim_routes}
	\vspace{-0.1in}
\end{figure}

\subsection{Visualization of Routing}
\label{ss:visualization}
\vspace{-0.05in}
We visualize the tensor activations and the gate values of each layer (see Fig. \ref{fig:1b} and Fig. \ref{fig:imtoim_routes}) by representing the average activation of each tensor (red) and gate value (blue) by their color intensities. 
The color intensity of tensor $j$ in layer $l$ represents the strength of prediction of $T^l_j$  normalized by the summation of the values of all tensors in the corresponding layer. Gates computed as softmax probabilities are mapped with the color intensity based on the corresponding probabilities. Blue colored lines between two adjacent layers illustrate the data flow from a particular tensor to the following layer and their width represents the weight associated with the data flow $g_{ij}$.

CAMNet being able to take different path combinations for inputs originating from different datasets is impressive. Figure \ref{fig:imtoim_routes} shows the different combinations of paths taken in CAMNet, along with the combination of activated gates for each image when the model is trained on the combined dataset. Taking different combinations of routes to build a network can be interpreted as an adaptive single model which switches between slightly different sub-modules. 

\subsection{Weight Histograms of Parallel Convolutions}
\label{ss:weights}
\begin{figure}[htbp]
	\begin{center}
		\begin{subfigure}[]{0.24\linewidth}
			\includegraphics[width=\linewidth]{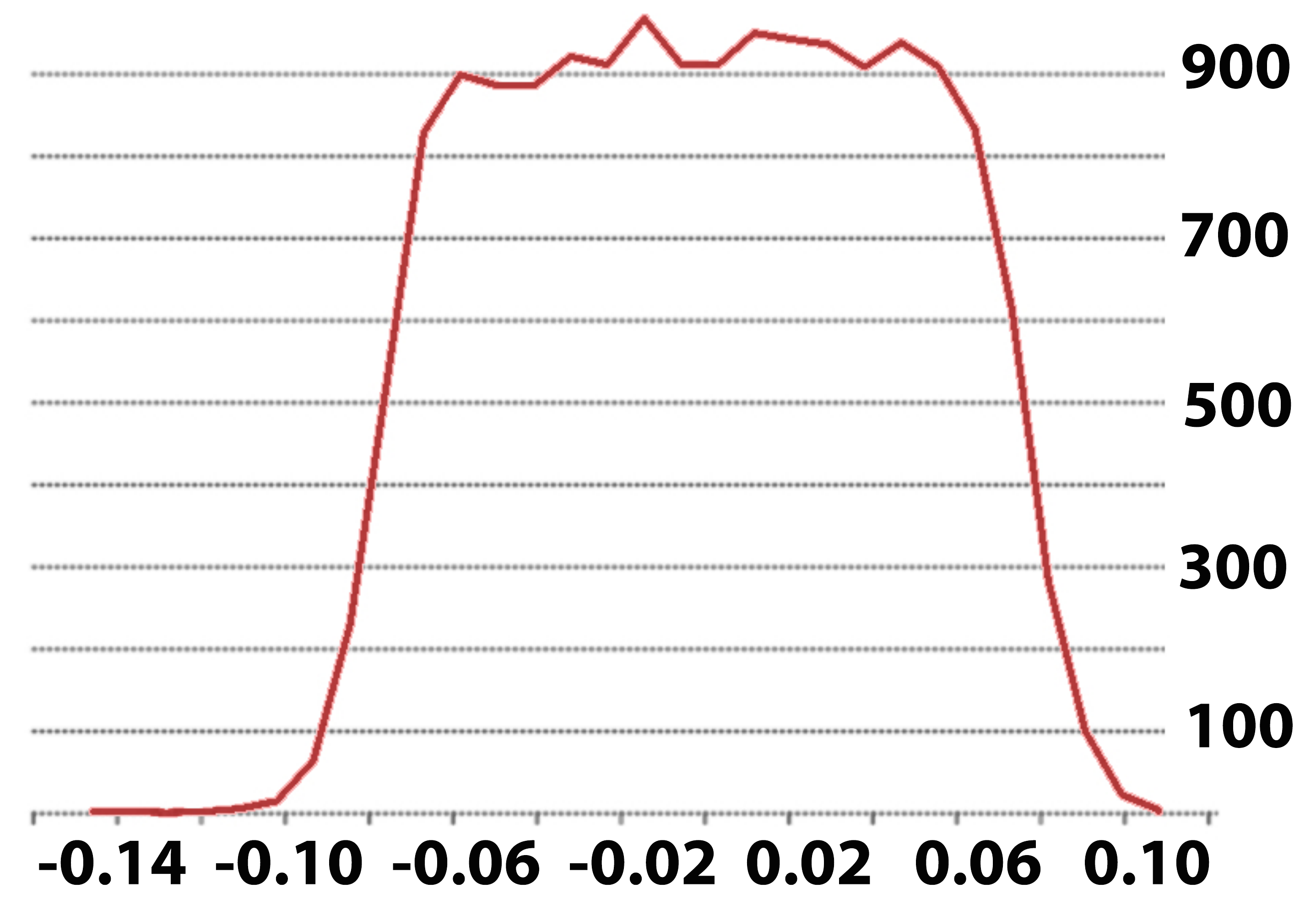} 
		\end{subfigure}
		\begin{subfigure}[]{0.24\linewidth}
			\includegraphics[width=\linewidth]{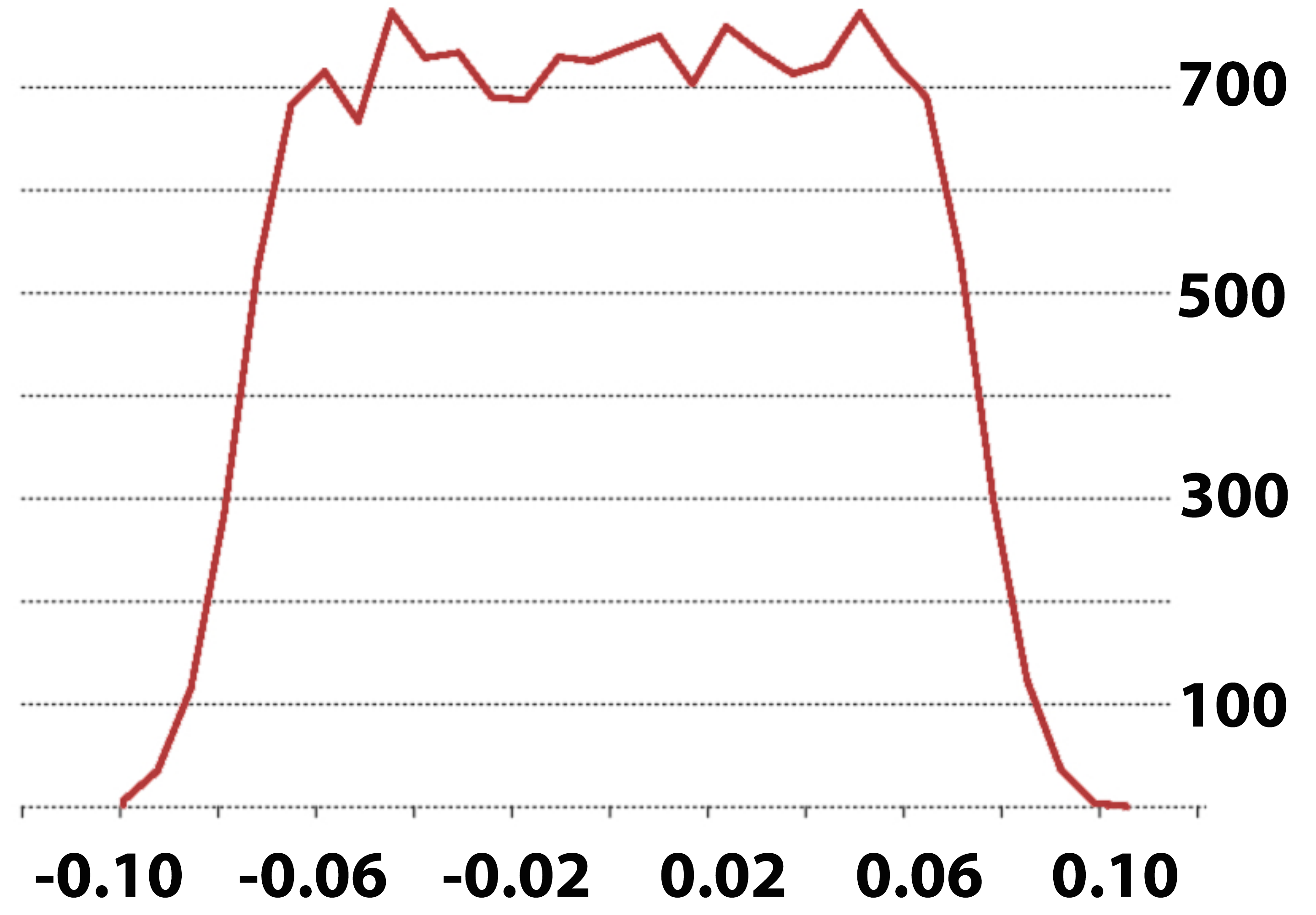}
		\end{subfigure}
		\begin{subfigure}[]{0.24\linewidth}
			\includegraphics[width=\linewidth]{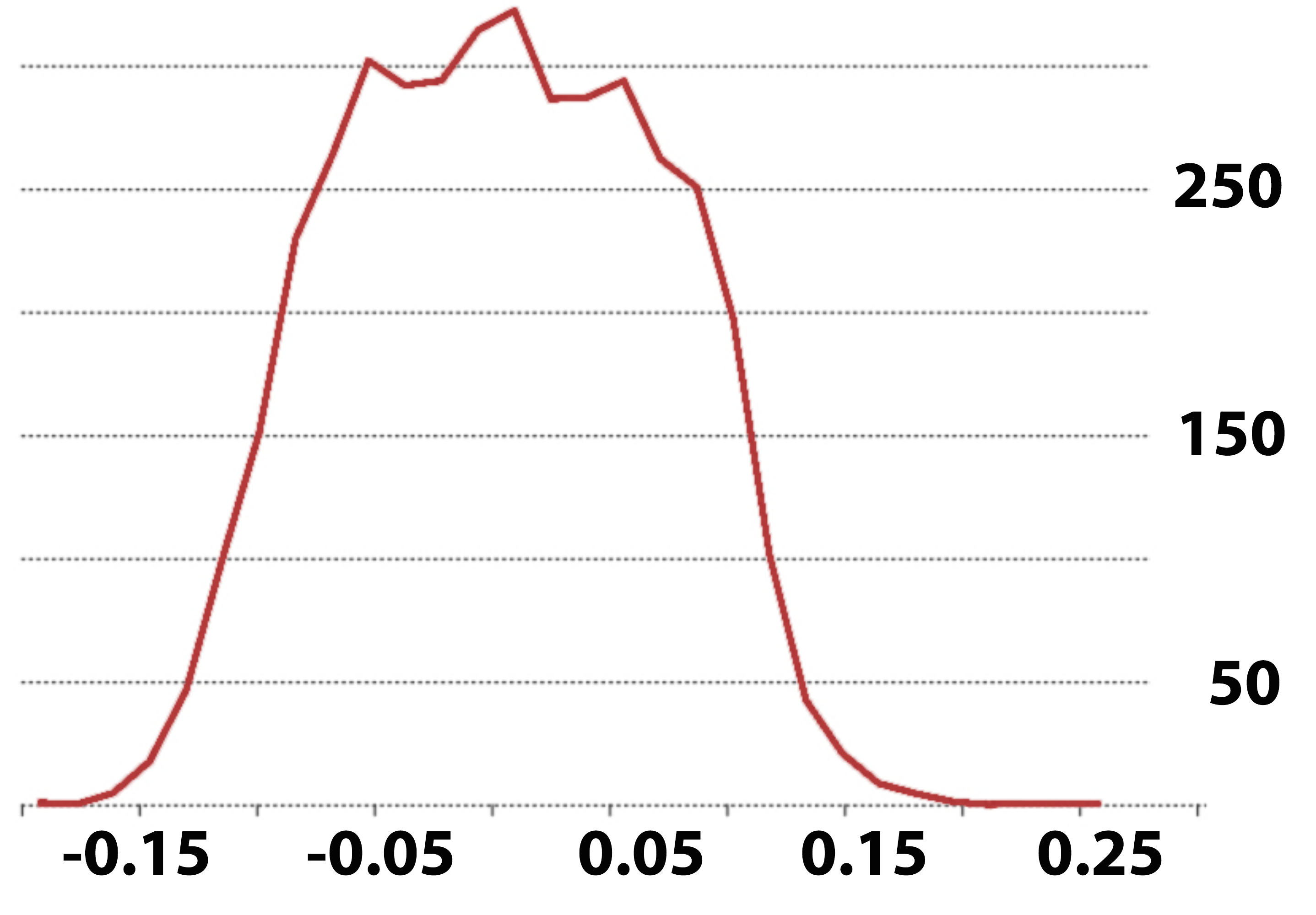}
		\end{subfigure}
		\begin{subfigure}[]{0.24\linewidth}
			\includegraphics[width=\linewidth]{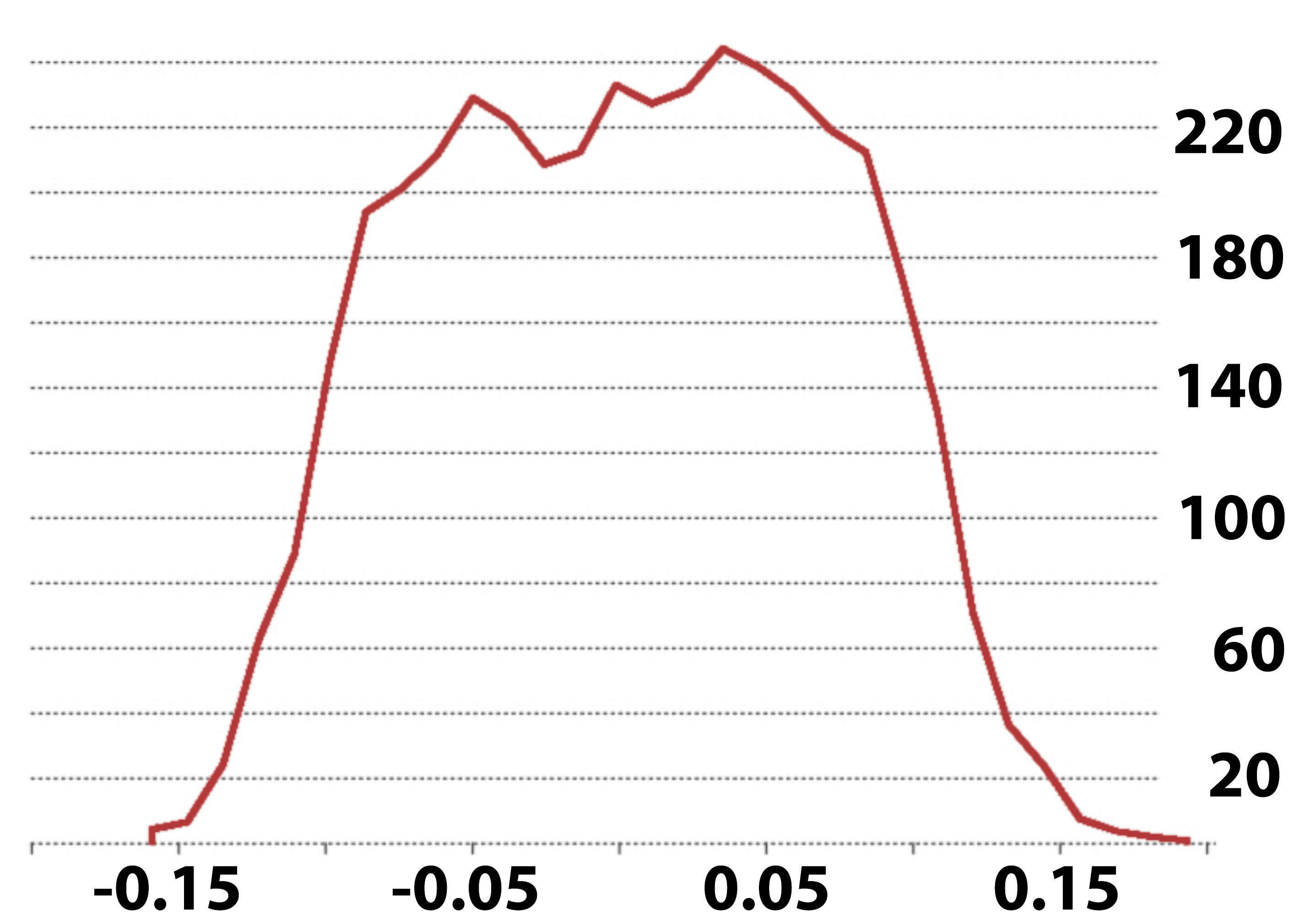}
		\end{subfigure}
		
		\begin{subfigure}[]{0.24\linewidth}
			\includegraphics[width=\linewidth]{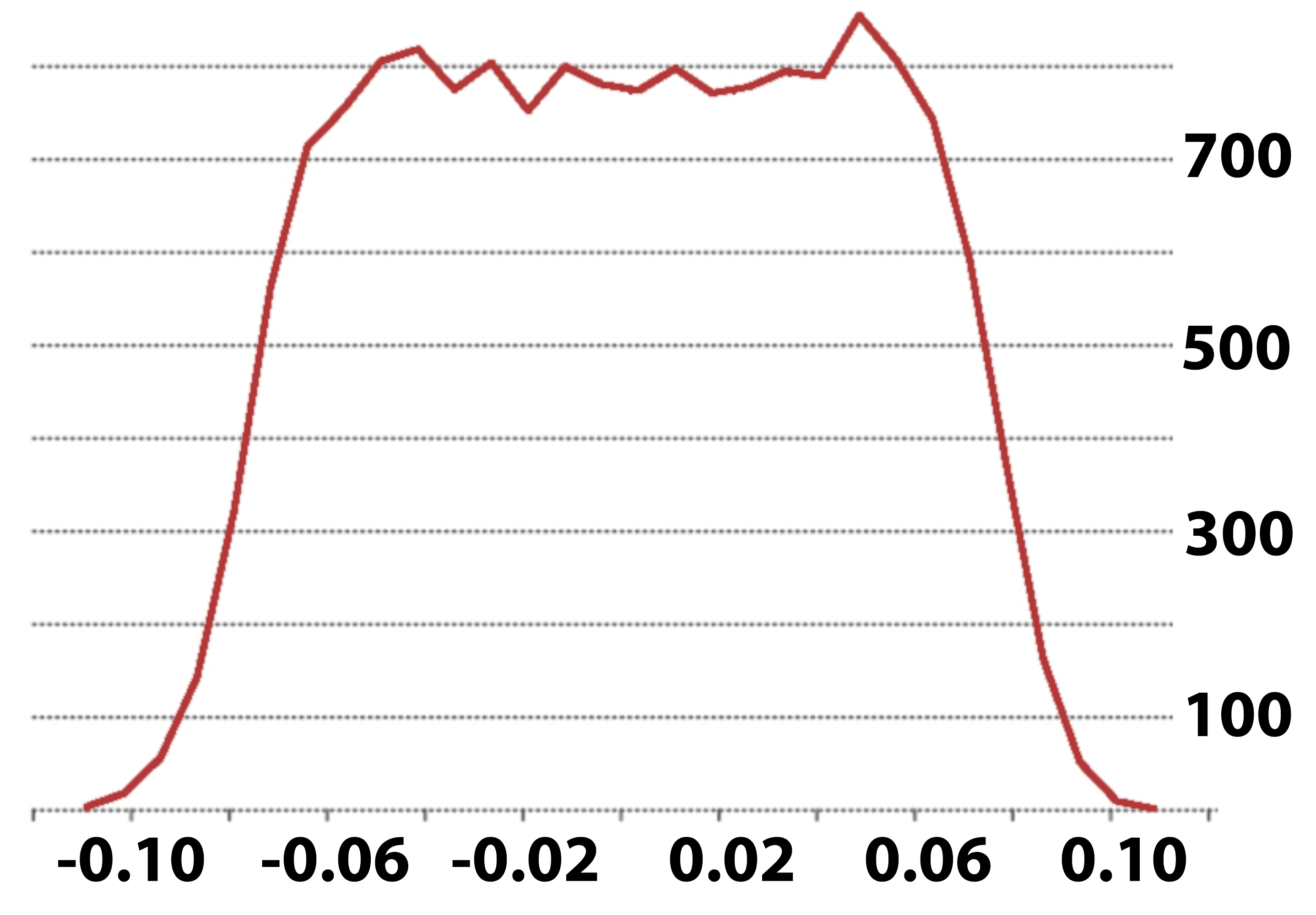}
			\caption{$Forw^2_1$} 
		\end{subfigure}
		\begin{subfigure}[]{0.24\linewidth}
			\includegraphics[width=\linewidth]{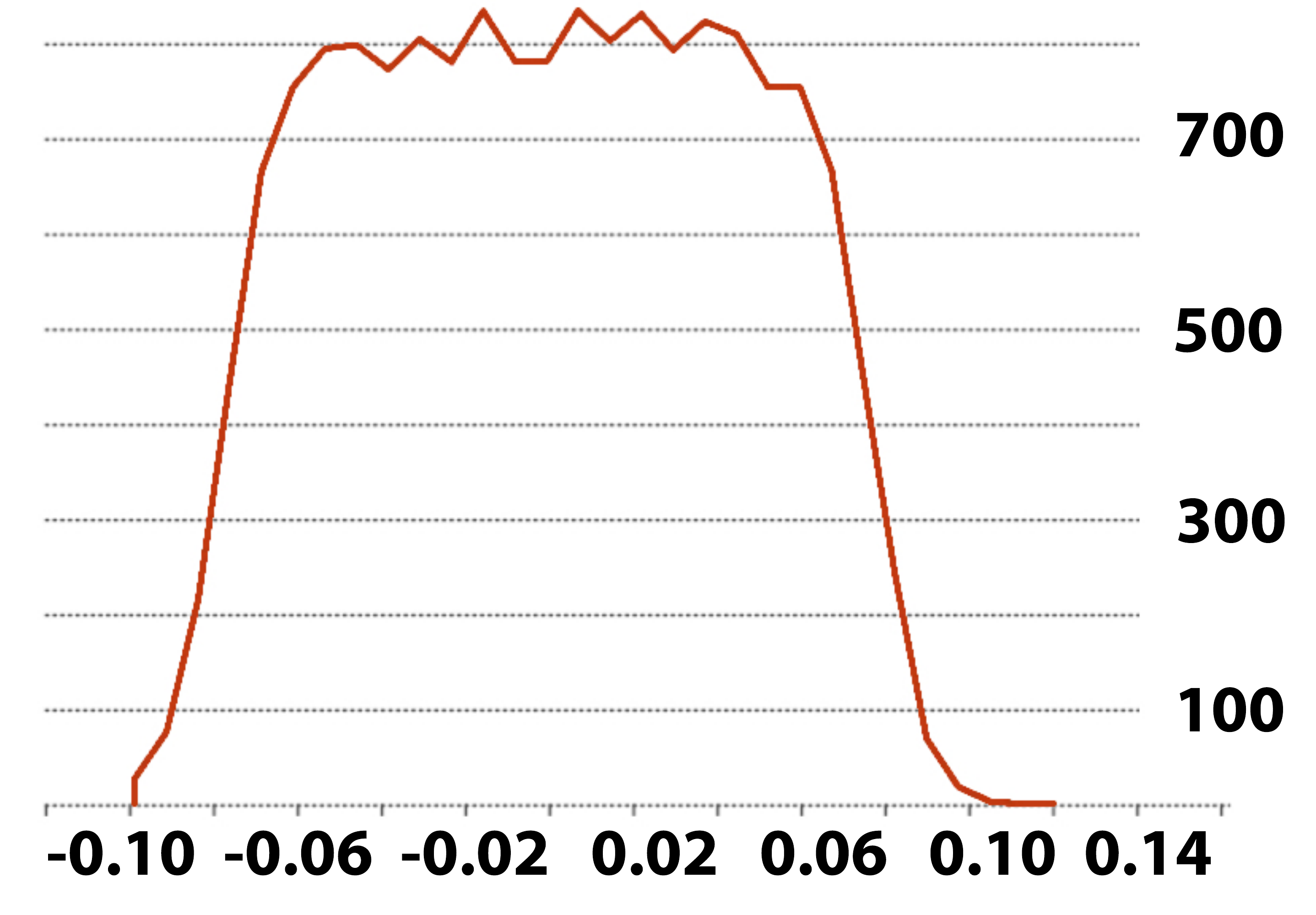}
			\caption{$Forw^2_2$}
		\end{subfigure}
		\begin{subfigure}[]{0.24\linewidth}
			\includegraphics[width=\linewidth]{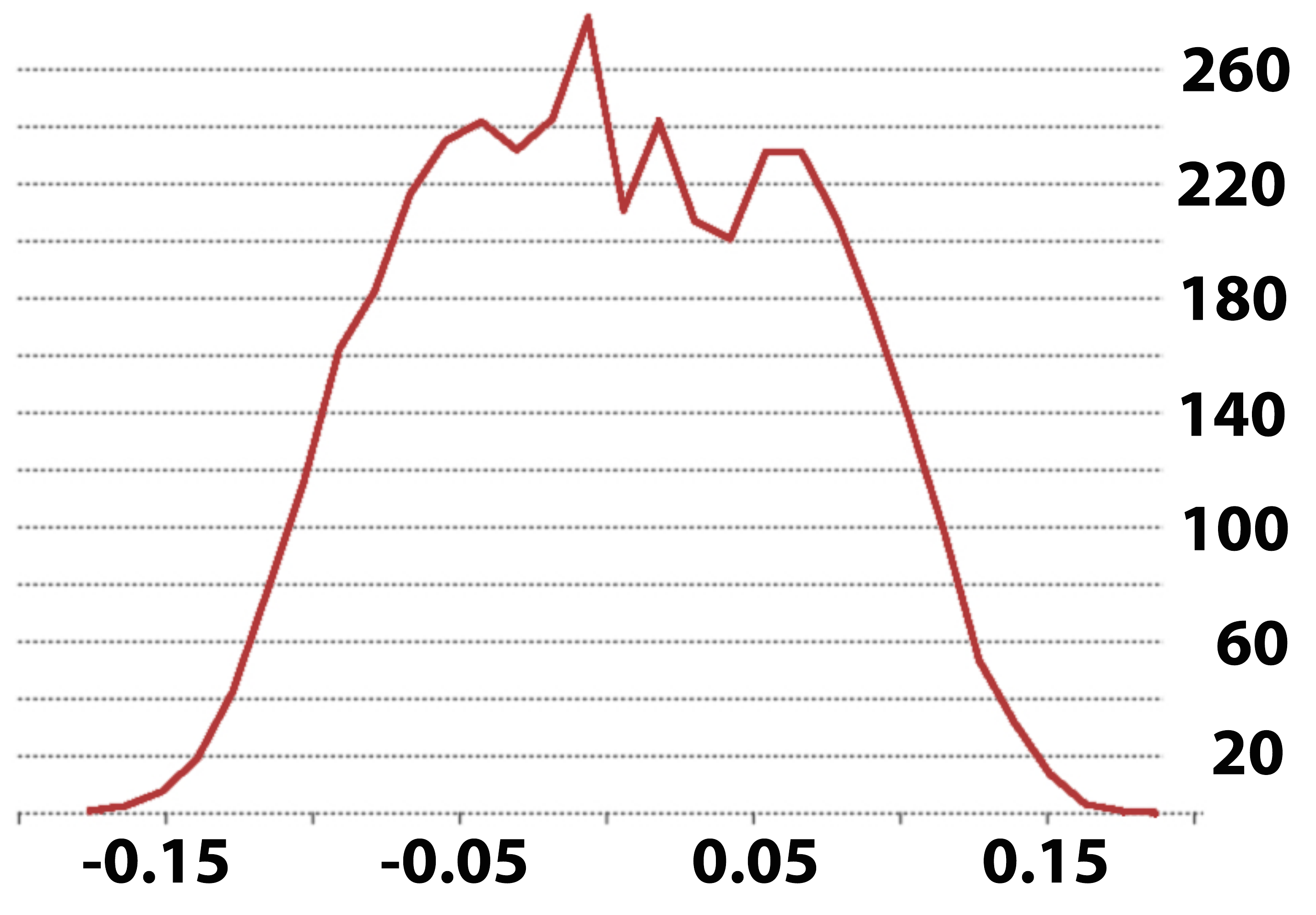}
			\caption{$Forw^9_1$}
		\end{subfigure}
		\begin{subfigure}[]{0.24\linewidth}
			\includegraphics[width=\linewidth]{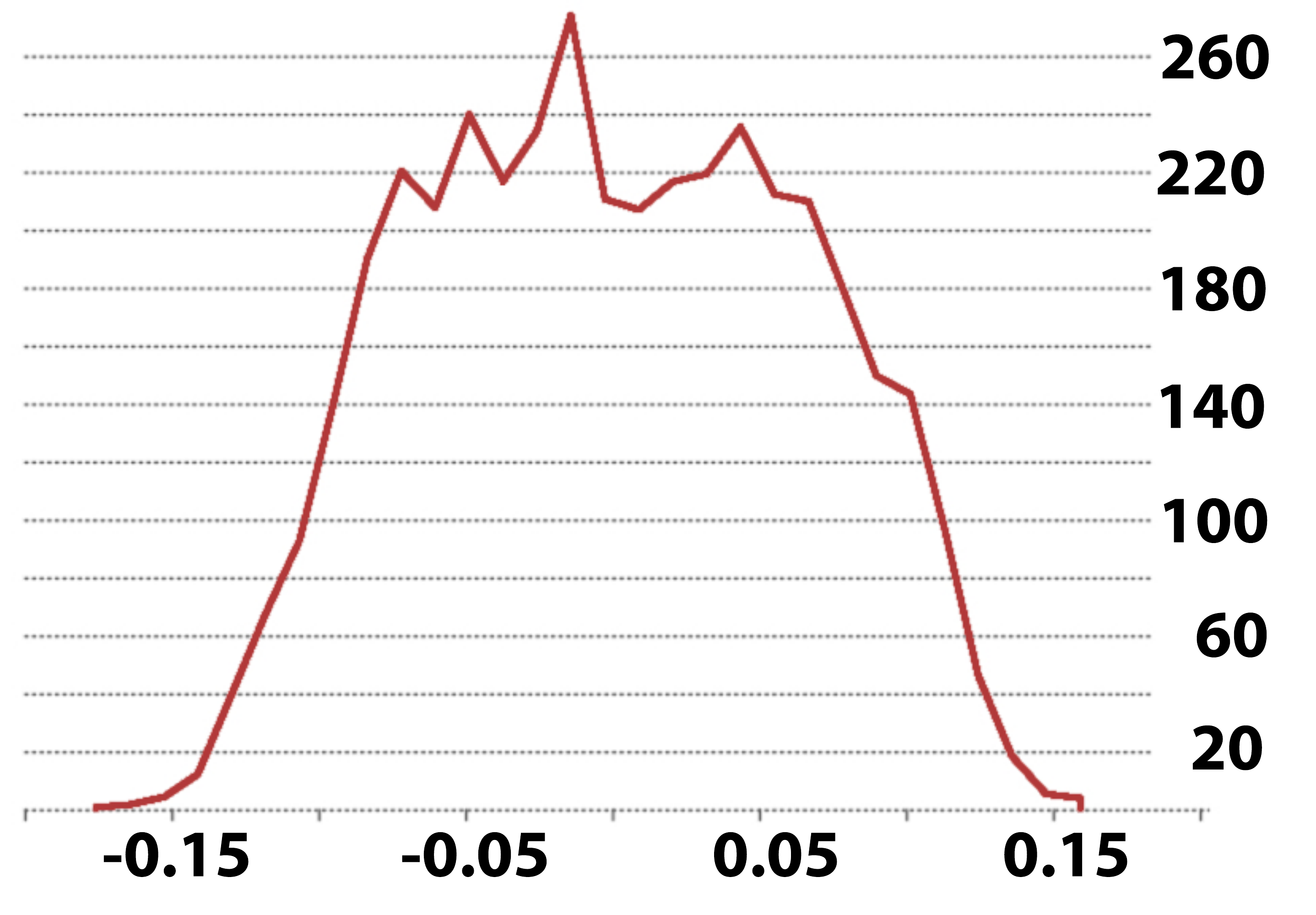}
			\caption{$Forw^9_2$}
		\end{subfigure}
	\end{center}
	\vspace{-0.2in}
	\caption{Weights Histograms of forward convolutions after $2^{nd}$ routing layer and $9^{th}$ routing layer. Top row shows CAMNet and bottom row shows similar multi-path network without gates. $Forw^l_i$ denotes the forward convolution carried by tensor $i$ at the output of routing layer $l$. Differences in histograms show that parallel paths have learned different portions of information in CAMNet.}
	\label{fig:histograms}
	\vspace{-0.1in}
\end{figure}

\vspace{-0.05in}
Here, we analyze the difference between convolutional weights learned across the parallel convolutions of particular layers. For this purpose, we compare the encoder-decoder version of CAMNet with a similar multi-path network and no routing in-between. Figure \ref{fig:histograms} shows the weight histograms of the forward layers (see Sec. \ref{ss:sa_forw_layers} for definition of forward layers) after $2^{nd}$ and $9^{th}$ routing layers. Since the CAMNet we test has two parallel tensors in each routing layer, we have two weight histograms of parallel forward convolutions per each layer. The figure illustrates that the histograms of parallel convolutions of an equivalent multi-path network with no routing are similar whereas the histograms of parallel convolutions of CAMNet with learnable routing are distinct. Such observations confirm that the parallel paths have learned different portions of information. 

\section{Discussion}
\label{se:discussion}
\vspace{-0.05in}
\subsection{Performance of CAMNet}
\label{ss:performance_camnet}
\vspace{-0.05in}
Our experiments show that the learnable routing between parallel tensors in CAMNet surpasses a similar multi-path network with no routing, a single-path similar-depth network, and even an equivalent deeper single-path network with similar number of parameters. The experiments further indicate that the improvement of CAMNet is not merely due to the number of parallel tensors, but due to the data-dependant routing algorithm which effectively uses  parallel resources. Increasing the width of CAMNet shows an improvement. However, for a given dataset, at some point increasing the width of CAMNet does not further improve the results.  

\subsection{Effect of Routing}
\label{ss:effect of routing}
\begin{figure}[ht]
	\begin{center}
		\begin{subfigure}[b]{\columnwidth}
			\includegraphics[width=\linewidth]{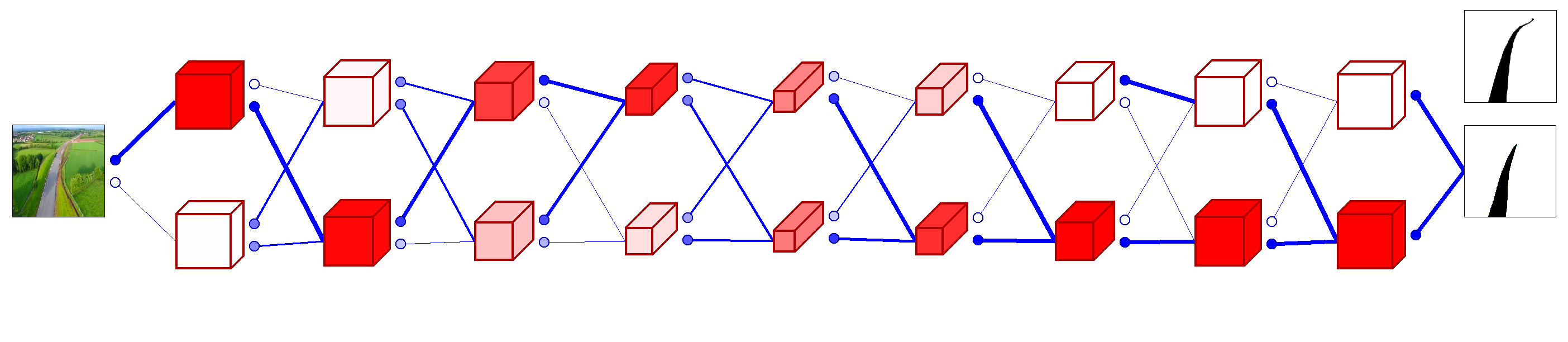}
		\end{subfigure}
		\begin{subfigure}[b]{\columnwidth}
			\includegraphics[width=\linewidth]{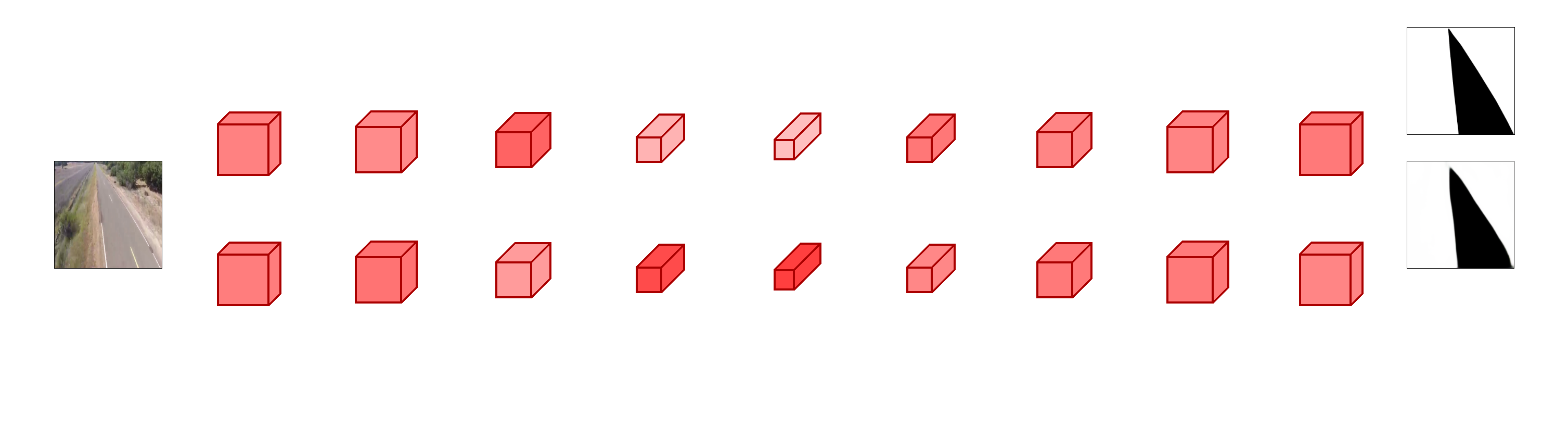}
		\end{subfigure}
	\end{center}
	\vspace{-0.2in}
	\caption{Top figure is CAMNet allocating resources based on the input, whereas the bottom figure shows a similar network without learnable routing, in which, there is no context-aware resource allocation. Hence, it shows a redundancy in parallel layers where similar features have been learned across the parallel tensors in the latter case.}
	\label{fig:withandwithout_routing}
	\vspace{-0.1in}
\end{figure}

\vspace{-0.05in}
We prefer CAMNet over multi-path CNN with no routing (MultiCNN) since it better utilizes a multi-path network. It creates a specific path for each input based on the underlying context related to the domain. Moreover, the network shares the common features of different contexts while allowing the network to learn more domain-specific features within the remaining tensors. Figure \ref{fig:withandwithout_routing} indicates the difference between information flow in CAMNet with a similar multi-path network without routing. 

One benefit of  data-dependant routing is that the parallel paths are able to extract relevant features for a particular input while not learning redundant features. The learned weights of parallel convolutions being distinct (Sec. \ref{ss:weights}) indicates that our network demotivates similar information being learned by parallel tensors. We can further observe the variation of learned information in parallel tensors by the network visualization for different input-output combinations in Figure \ref{fig:imtoim_routes}. Figure \ref{fig:withandwithout_routing} further illustrates the effect of routing between tensors where a similar multi-path network with no routing does not have the regulation of flow and allocation of resources depending on the context.

\subsection{Number of Parallel Tensors}
\label{ss:parallel_tensors}
\vspace{-0.05in}
The number of parallel tensors included in each layer depends on the diversity of the domain and task. For a given dataset or a set of datasets, the performance of CAMNet increases with the number of parallel tensors and saturates at some point. The performance in individual datasets seems to saturate at CAMNet3; CAMNet4 does not provide better results than CAMNet3 in most individual datasets (Table \ref{tab:classi}). Although CAMNet4 shows the best results even in lifelong learning and joint training, the increment of accuracy is very low compared to the increment from CAMNet2 to CAMNet3.(Table \ref{tab:classi} and Figure \ref{fig:lwf_plots})

It is further possible to have different number of parallel tensors in each routing layer. We do not investigate this fact in this paper. However, depending on the degree of similarity between datasets or within a dataset, and the difference between intended outputs, the exact number of parallel tensors in each layer will vary.



\section{Conclusion}
\label{se:conclusion}
\vspace{-0.05in}
In this paper, we presented CAMNet, a multi-path neural network with data-dependant routing for context awareness. We evaluated classification version of CAMNet architecture with classification datasets and encoder-decoder version of CAMNet with pixel labeling. We show that CAMNet outperforms equivalent baseline single path and multi-path networks and even deeper single-path networks when tested with both individual datasets and dataset combinations. In addition, CAMNet surpasses baseline networks in lifelong learning, preserving already learned information better. CAMNet can address a wider context awareness which enable it to learn from even multiple different contexts better than comparable deep networks. The data-dependant routing enables the model to wisely allocate resources, deciding between domain-specific and shared tensors, learning the information flow end-to-end. We believe that the idea of partially utilizing a network with the help of learnable routing is an impressive solution to context-awareness and multi-domain learning.

{\small
\balance
\bibliographystyle{ieee}

}
\end{document}